\documentclass[11pt,a4paper]{article}
\usepackage{times,latexsym}
\usepackage{url}
\usepackage[T1]{fontenc}

    \usepackage[acceptedWithA]{tacl2018v2}
\usepackage[]{tacl2018v2}

\usepackage{xspace,mfirstuc,tabulary}

\newif\iftaclinstructions
\taclinstructionsfalse 
\iftaclinstructions
\renewcommand{\confidential}{}
\renewcommand{\anonsubtext}{(No author info supplied here, for consistency with
TACL-submission anonymization requirements)}
\newcommand{\instr}
\fi

\iftaclpubformat 

\else

\fi

\title{Quantifying Social Biases in NLP:\\A Generalization and Empirical Comparison\\of Extrinsic Fairness Metrics}

\author{
 Paula Czarnowska$^\spadesuit$\\
 University of Cambridge \\
 {\sf pjc211@cam.ac.uk} \And
  Yogarshi Vyas \\
 Amazon AI \\
 {\sf yogarshi@amazon.com } 
 \And
  Kashif Shah \\
 Amazon AI \\
 {\sf shahkas@amazon.com } 
}

\usepackage{amssymb}
\usepackage{amsthm}
\usepackage{color}
\usepackage{enumitem}
\usepackage{amsmath}
\usepackage{graphicx}
\usepackage{url}
\usepackage{comment}
\usepackage{diagbox}
\usepackage{caption}
\usepackage{verbatim}
\usepackage{mathrsfs}
\usepackage{verbatim}
\usepackage{times}
\usepackage{url}
\usepackage{latexsym}
\usepackage{color}
\usepackage{float}
\usepackage{array}
\usepackage{amsmath,amsfonts,amssymb}
\usepackage{algorithm}
\usepackage{algpseudocode}
\usepackage{graphicx}
\usepackage{booktabs}
\usepackage{pifont}
\usepackage{fancyvrb}
\usepackage{footnote}
\usepackage[capitalise]{cleveref}
\usepackage{multirow}
\usepackage{makecell}
\usepackage[normalem]{ulem}

\usepackage{tikz}

\usepackage{array}

\usepackage{caption}
\usepackage{subcaption}
\usepackage[toc,page]{appendix}
\usepackage{csvsimple}
\usepackage{longtable}
\usepackage{supertabular}
\usepackage{color, colortbl}

\usepackage{background}
\usepackage{lipsum}

\backgroundsetup{contents={Pre-MIT Press publication version}, placement=top, scale=0.9, color=black, vshift=-0.5cm, hshift=8.3cm}

\newcolumntype{L}[1]{>{\raggedright\let\newline\\\arraybackslash\hspace{0pt}}m{#1}}
\newcolumntype{C}[1]{>{\centering\let\newline\\\arraybackslash\hspace{0pt}}m{#1}}
\newcolumntype{R}[1]{>{\raggedleft\let\newline\\\arraybackslash\hspace{0pt}}m{#1}}

\usepackage{stmaryrd}

\newcolumntype{P}[1]{>{\centering\arraybackslash}p{#1}}
\definecolor{ghostwhite}{rgb}{0.95, 0.95, 0.96}
\definecolor{ghostgray}{rgb}{0.90, 0.90, 0.90}

\newcommand{\mnote}[1]
{
}

\definecolor{othercolor}{rgb}{0.1, 0.35, 0.35}

\newcommand{\emphh}[1]{\textit{#1}}
\newcommand{\word}[1]{\textit{#1}}
\newcommand{\defn}[1]{\textit{#1}}

\newcommand*\circled[1]{\tikz[baseline=(char.base)]{
            \node[shape=circle,draw,minimum size=0.35cm,inner sep=1pt] (char) {\tiny #1};}}

\newcommand*\ccircled[1]{\tikz[baseline=(char.base)]{
            \node[shape=circle,draw,minimum size=0.35cm,inner sep=1pt, fill=ghostgray] (char) {\tiny #1};}}

\setlist{noitemsep}
\newcommand{\ignore}[1]{}

\newcommand\blfootnote[1]{%
  \begingroup
  \renewcommand\thefootnote{}\footnote{#1}%
  \addtocounter{footnote}{-1}%
  \endgroup
}

\definecolor{palered}{rgb}{0.95, 0.9, 0.9}
\definecolor{palegreen}{rgb}{0.9, 0.95, 0.9}

\date{}

\begin{document}
\maketitle
\begin{abstract}

Measuring bias is key for better understanding and addressing unfairness in NLP/ML models. This is often done via {fairness metrics} which quantify the differences in a model's behaviour across a range of demographic groups. In this work, we shed more light on the differences and similarities between the fairness metrics used in NLP. First, we unify a broad range of existing metrics under three \textbf{generalized fairness metrics}, revealing the connections between them. Next, we carry out an extensive empirical comparison of existing metrics and demonstrate that the observed differences in bias measurement can be systematically explained via differences in parameter choices for our generalized metrics. 

\end{abstract}

\section{Introduction}
\blfootnote{\noindent $^\spadesuit$ Work done during an internship at Amazon AI.} 
The prevalence of unintended social biases in NLP models has been recently identified as a major concern for the field. A number of papers have published evidence of uneven treatment of different demographics  \cite{dixon-2018, zhao-etal-2018-gender, rudinger-etal-2018-gender, garg-2019, borkan_nuanced_2019, stanovsky-etal-2019-evaluating, gonen-webster-2020-automatically, huang-etal-2020-reducing, nangia-etal-2020-crows}, which can reportedly cause a variety of serious harms, like unfair allocation of opportunities or unfavorable representation of particular social groups \cite{blodgett_language_2020}.

Measuring bias in NLP models is key for better understanding and addressing unfairness. This is often done via \textbf{fairness metrics} which quantify the differences in a model's behaviour across a range of social groups. 
The community has proposed a multitude of such metrics \cite{dixon-2018, garg-2019, huang-etal-2020-reducing, borkan_nuanced_2019, gaut-etal-2020-towards}. In this paper, we aim to shed more light on how those varied means of quantifying bias differ and what facets of bias they capture. Developing such understanding is crucial for drawing reliable conclusions and actionable recommendations regarding bias. We focus on bias measurement for downstream tasks as \citet{goldfarb2020intrinsic} have recently shown that there is no reliable correlation between bias measured intrinsically on, for example, word embeddings, and bias measured extrinsically on a downstream task. We narrow down the scope of this paper to tasks which do not involve prediction of a sensitive attribute.

We survey 146 papers on social bias in NLP and unify the multitude of disparate metrics we find under three \textbf{generalized fairness metrics}. Through this unification we reveal the key connections between a wide range of existing metrics---we show that they are simply \emphh{different parametrizations} of our generalized metrics. Next, we empirically investigate the role of different metrics in detecting the systemic differences in performance for different demographic groups, i.e., differences in \word{quality of service} \cite{jacobs-2020}. We experiment on three transformer-based models---two models for sentiment analysis and one for named entity recognition (NER)---which we evaluate for fairness with respect to seven different sensitive attributes, qualified for protection under the United States federal anti-discrimination law:\footnote{\url{https://www.ftc.gov/site-information/no-fear-act/protections-against-discrimination}} \word{Gender}, \word{Sexual Orientation}, \word{Religion}, \word{Nationality}, \word{Race}, \word{Age} and \word{Disability}. {Our results highlight the differences in bias measurements across the metrics and we discuss how these variations can be systematically explained via different parameter choices of our generalized metrics.} 
Our proposed unification and observations can guide decisions about which metrics (and parameters) to use, allowing researchers to focus on the pressing matter of bias mitigation, rather than reinventing parametric variants of the same metrics. While we focus our experiments on English, the metrics we study are language-agnostic and our methodology can be trivially applied to other languages. 

We release our code with implementations of all metrics discussed in this paper.\footnote{We will provide the link in the MIT Press version.} Our implementation mirrors our generalized formulation (\cref{gen-metrics}), which simplifies the creation of new metrics. We build our code on top of \textsc{CheckList}\footnote{https://github.com/marcotcr/checklist} \cite{ribeiro_beyond_2020}, making it compatible with the \textsc{CheckList} testing functionalities; i.e., one can evaluate the model using the fairness metrics, as well as the \textsc{CheckList}-style tests, like \defn{invariance}, under a single \textbf{bias evaluation framework}.

\vspace{-0.1cm}
\section{Background}

\subsection{Terminology}
We use the term \textbf{sensitive attribute} to refer to a category by which people are qualified for protection, e.g., \word{Religion} or \word{Gender}. For each {sensitive attribute} we define a set of \textbf{protected groups} $T$, e.g., for \word{Gender}, $T$ could be set to $\{\text{female, male, non-binary}\}$. Next, each protected group can be expressed through one of its \textbf{identity terms}, $I$; e.g., for the protected group \word{female} those terms {could} be \{woman, female, girl\} or a set of typically female names. 

\subsection{Definitions of Fairness in NLP}\label{sec:definitions}
The metrics proposed to quantify bias in NLP models across a range of social groups can be categorized based on whether they operationalize notions of {group} or {counterfactual} fairness. In this section we give a brief overview of both and encourage the reader to consult \citet{hutchinson201950} for a broader scope of literature on fairness, dating back to the 1960s.

\vspace{-0.1cm}
\paragraph{Group fairness} requires parity of some statistical measure across a small set of protected groups \cite{chouldechova2018frontiers}.
Some prominent examples are \word{demographic parity} \cite{dwork-2012}, which requires equal positive classification rate across different groups,
or \word{equalized odds} \cite{hardt-2016} which for binary classification requires equal true positive and false negative rates. In NLP, group fairness metrics are based on performance comparisons for {different} sets of examples, e.g., the comparison of two F1 scores: one for {examples mentioning female names and one for examples with male names.}

\begin{table}
    \centering
    \footnotesize
    \begin{tabular}{lll}
        \toprule
        \makecell{Source\\Example}  & Female & Male \\
        \midrule 
          \multirow{3}{*}{\makecell{I like\\\{person\}.}} & I like Anna. & I like Adam.\\
         & I like Mary. & I like Mark. \\
         & I like Liz. & I like Chris.  \\
        \midrule 
        \multirow{3}{*}{\makecell{\{Person\}\\has friends.}} & Anna has friends. & Adam has friends. \\
        &  Mary has friends. & Mark has friends. \\
        & Liz has friends. & Chris has friends. \\
        \bottomrule
    \end{tabular}
    \caption{Example of counterfactual fairness data. $T = \{\text{female, male}\}$ and $|I| = 3$ for both groups.} \label{tab:examples}
\end{table}

\paragraph{Counterfactual fairness} requires parity {for two or more versions of an individual, one from the actual world and others from counterfactual worlds} in which the individual belongs to a \emphh{different protected group}; i.e., {it requires invariance to the change} of the protected group \cite{kusner-2017}. Counterfactual fairness is often viewed as a type of {individual fairness}, which asks for similar individuals to be treated similarly \cite{dwork-2012}. In NLP, counterfactual fairness metrics are based on comparisons of performance for variations \emphh{of the same sentence}, which differ in mentioned identity terms. 
Such data can be created through perturbing real-world sentences or creating synthetic sentences from templates.

In this work, we require that {for each protected group} there exists \emphh{at least one} sentence variation for every source example (pre-perturbation sentence or a template). In practice, the number of variations for each protected group will depend on the cardinality of $I$ {(\cref{tab:examples})}.
In contrast to most NLP works \cite{dixon-2018, garg-2019, sheng-etal-2020-towards}, we allow for a protected group to be realized as more than one identity term.
To allow for this, we separate the variations for each source example into $|T|$ \emphh{sets}, each of which can be viewed as a separate counterfactual world.

\section{Generalized Fairness Metrics} \label{gen-metrics}

We introduce three \textbf{generalized fairness metrics} which are based on different comparisons between protected groups and are model and task agnostic. They are defined in terms of two parameters:
\begin{enumerate}
    \item[(i)] 
    A scoring function, $\phi$, which calculates the \defn{score} on a subset of examples. The \defn{score} is a base measurement used to calculate the metric and can be either a scalar or a set (see \cref{tab:metrics-table} for examples).
    \vspace{0.1cm}
    \item[(ii)] 
    A comparison function, $d$, which takes a range of different scores---computed for different subsets of examples---and outputs a single scalar value.
\end{enumerate}

\noindent {Each of the three metrics is conceptually different and is most suitable in different scenarios; the choice of the most appropriate one depends on the scientific question being asked. Through different choices for $\phi$ and $d$, we can systematically formulate a broad range of different fairness metrics, targeting different types of questions. We demonstrate this in  \cref{sec:existing_metrics} and \cref{tab:metrics-table}, where we show that many metrics from the NLP literature can be viewed as parametrizations of the metrics we propose here. }
\noindent To account for the differences between group and counterfactual fairness (\cref{sec:definitions}) we define \emphh{two different versions of each metric}.

\paragraph{Notation}
Let $T = \{t_1, t_2,\ ..., t_{|T|}\}$ be a set of all protected groups for a given sensitive attribute, e.g., \word{Gender}, and $\phi(A)$ be the \defn{score} for some set of examples $A$. This score can be either a set or a scalar, depending on the parametrization of $\phi$. For group fairness, let $S$ be the set of all evaluation examples. We denote a subset of examples associated with a protected group $t_{i}$ as $S^{t_{i}}$. For counterfactual fairness, let $X = \{x_1, x_2, ..., x_{|X|}\}$ be a set of \defn{source examples}, e.g., sentences pre-perturbation, and ${S'} = \{{S'}_{1}, {S'}_{2}, ..., {S'}_{|S|}\}$ be a \emphh{set of sets} of {evaluation examples}, where ${S'}_{j}$ is a set of all variations of a source example $x_{j}$, i.e., there is a one-to-one correspondence between ${S'}$ and $X$. We use ${S'}_{j}^{t_i}$ to denote a subset of ${S'}_{j}$ associated with a protected group $t_{i}$. For example, if $T = \{$\text{female, male}$\}$ and the templates were defined as in \cref{tab:examples}, then ${S'}_{1}^{\text{female}} = \{$\text{`I like Anna.', `I like Mary.', `I like Liz.'}$\}$.

\subsection{Pairwise Comparison Metric}

{Pairwise Comparison Metric (PCM)} quantifies how distant, on average, the scores for two different, randomly selected groups are. It is suitable for examining whether and to what extent the chosen protected groups differ from one another. For example, for the sensitive attribute \word{Disability}, are there any performance differences for cognitive vs mobility vs {no disability}? We define Group (\ref{gpcm}) and Counterfactual (\ref{cpcm}) PCM as follows:
\begin{equation}\label{gpcm}
\frac{1}{N}\, \sum_{t_i, t_j \in {T \choose 2} }\, d\left(\phi(S^{t_i}), \phi(S^{t_j}) \right)
\end{equation}
\vspace{-0.25cm}
\begin{equation}\label{cpcm}
\frac{1}{|{S'}|\; N} \sum_{{S'}_j \in {S'}} \sum_{t_i, t_k \in {T \choose 2} }\, d\left(\phi({S'}_j^{t_i}), \phi({S'}_j^{t_k}) \right)
\end{equation}

\noindent where $N$ is a normalizing factor, e.g., ${|T| \choose 2}$. 

\subsection{Background Comparison Metric} 

{Background Comparison Metric (BCM)} relies on a comparison between the score for a protected group and the score of its \textbf{background}. The definition of the background depends on the task at hand and the investigated question. For example, if the aim is to answer whether the performance of a model for the group differs from the model's \textit{general} performance, the background can be a set of \emphh{all} evaluation examples.
{Alternatively, if the question of interest is whether the groups considered disadvantaged are treated differently than some privileged group, the background can be a set of examples associated with that privileged group. {In such a case, $T$ should be narrowed down to the disadvantaged groups only.}}
For counterfactual fairness the background could be the unperturbed example, allowing us to answer whether a model's behaviour differs for any of the counterfactual versions of the world. Formally, we define Group (\ref{gbcm}) and Counterfactual (\ref{cbcm}) BCM as follows:

\begin{equation}\label{gbcm}
 \frac{1}{N} \sum_{t_i \in T}\, d\left(\phi(\beta^{t_i, S}), \phi(S^{t_i}) \right)
\end{equation}
\vspace{-0.2cm}
\begin{equation}\label{cbcm}
\frac{1}{|{S'}|\, N} \sum_{{S'}_j \in {S'}} \sum_{t_i \in T}\, d\left(\phi(\beta^{t_i, {S'}_j}), \phi({S'}_j^{t_i}) \right)
\end{equation}

\noindent where $N$ is a normalizing factor and $\beta^{t_i, S}$ is the background for group $t_i$ for the set of examples $S$. 
 
\paragraph{Vector-valued BCM}
In its basic form BCM aggregates the results obtained for different protected groups in order to return a single scalar value. Such aggregation provides a concise signal about the presence and magnitude of bias, but it does so at the cost of losing information. Often, it is important to understand how different protected groups contribute to the resulting outcome. This requires the individual group results not to be accumulated; i.e., dropping the $\frac{1}{N}\sum_{t_i \in T}\, $ term from equations \ref{gbcm} and \ref{cbcm}. We call this version of BCM, the {vector-valued BCM (VBCM)}.

\subsection{Multi-group Comparison Metric} 

{Multi-group Comparison Metric (MCM)} differs from the other two in that the comparison function $d$ takes as arguments the scores for \emphh{all protected groups}. This metric can quantify the global effect that a sensitive attribute has on a model's performance{; e.g., whether the change of \word{Gender} has any effect on model's scores. It can provide a useful initial insight, but further inspection is required to develop better understanding of the underlying bias, if it is detected.}\\
Group (\ref{gmcm}) and Counterfactual (\ref{cmcm}) MCM are defined as:

\begin{equation}\label{gmcm}
d(\phi(S^{t_1}), \phi(S^{t_2}), ..., \phi(S^{t_{|T|}}))
\end{equation}
\vspace{-0.35cm}
\begin{equation}\label{cmcm}
\frac{1}{|{S'}|} \sum_{{S'}_j \in {S'}} d(\phi({S'}_j^{t_1}), \phi({S'}_j^{t_2}), ..., \phi({S'}_j{t_{|T|}}))
\end{equation}

\begin{table*}[!t]
    \centering
    \footnotesize
    \renewcommand{\arraystretch}{1.2}
    \begin{tabular}{p{0.2cm} l c  l  c  c  c  }
    \toprule
        & Metric & \makecell{Gen.\\Metric} & $\phi(A)$ & $d$ & $N$ & $\beta^{t_i, S}$  \\
    \midrule        
        \multicolumn{7}{c}{\multirow{1}{*}{\textsc{Group metrics}}}\\
    \midrule        
        \circled{1} 
            & \makecell[l]{False Positive Equality\\Difference (FPED)}
            & \multirow{4}{*}{BCM} & {False Positive Rate}
            & \makecell{$| x - y |$} & 1 & $S$ \\
        \circled{2} 
            & \makecell[l]{False Negative Equality\\Difference (FNED)}
            & & {False Negative Rate}
            & \makecell{$| x - y |$}& 1 & $S$ \\
        \circled{3}
            & \makecell[l]{Average Group\\Fairness (AvgGF)}
            &  & \makecell[l]{$\{f(x, 1) \mid x \in A\}$}
            & $W_1(X, Y)$  & $|T|$ & $S$  \\
    \midrule 
        \circled{4}
            & \makecell[l]{FPR Ratio}
            & \multirow{4}{*}{VBCM}  & False Positive Rate
            & \makecell{\large $\frac{y}{x}$} & - & $S \setminus S^{t_i}$\\ 
        \circled{5}
            & \makecell[l]{Positive Average\\Equality Gap (PosAvgEG)}
            &   & \makecell[l]{$\{f(x, 1) \mid x \in A, y(x) = 1\}$}
            & \makecell{$\frac{1}{2} - \frac{MWU(X, Y)}{|X||Y|}$}  & - & $S \setminus S^{t_i}$ \\
        \circled{6}
            &\makecell[l]{Negative Average\\Equality Gap (NegAvgEG)}
            &   & \makecell[l]{$\{f(x, 1) \mid x \in A, y(x) = 0\}$}
            & \makecell{$\frac{1}{2} - \frac{MWU(X, Y)}{|X||Y|}$}  & - & $S \setminus S^{t_i}$ \\
    \midrule 
        \circled{7}  
            & \makecell[l]{Disparity Score}
            & \multirow{10}{*}{PCM}  & F1
            & \makecell{$| x - y |$}& $|T|$ & - \\
        \circled{8} 
            & \makecell[l]{*TPR Gap}
            &  & True Positive Rate
            & \makecell{$| x - y |$}& ${|T| \choose 2}$ & - \\
        \circled{9}  
            & \makecell[l]{*TNR Gap}
            &  & True Negative Rate
            & \makecell{$| x - y |$} & ${|T| \choose 2}$ & - \\
        \circled{10}
            & \makecell[l]{*Parity Gap}
            &  & {\large $\frac{|\{ x \mid x \in A, \hat{y}(x) = y(x)\}|}{|A|}$}
            & \makecell{$| x - y |$} & ${|T| \choose 2}$ & - \\
        \ccircled{11}
            & \makecell[l]{*Accuracy Difference}
            &  & {Accuracy}
            & \makecell{$x - y$} & 1 & - \\      
        \ccircled{12}
            & \makecell[l]{*TPR Difference}
            &  & {True Positive Rate}
            & \makecell{$x - y$} & 1 & - \\      
        \ccircled{13}
            & \makecell[l]{*F1 Difference}
            &  & {F1}
            & \makecell{$x - y$} & 1 & - \\    
        \ccircled{14}
            & \makecell[l]{*LAS Difference}
            &  & {LAS}
            & \makecell{$x - y$} & 1 & - \\  
        \ccircled{15}
            & \makecell[l]{*Recall Difference}
            &  & {Recall}
            & \makecell{$x - y$} & 1 & - \\  
        \ccircled{16}
            & \makecell[l]{*F1 Ratio}
            &  & {Recall}
            & \makecell{\large{$\frac{x}{y}$}} & 1 & - \\  
    \midrule        
      \multicolumn{7}{c}{\multirow{1}{*}{\textsc{Counterfactual metrics}}}\\
    \midrule  
        \circled{17} 
            & \makecell[l]{Counterfactual Token\\Fairness Gap (CFGap)}
            & \multirow{1}{*}{BCM} & \makecell[l]{$f(x, 1),\: A = \{x\}$}
            & \makecell{$| x - y |$}& ${|T|}$ & \makecell{$\{x_j\}$} \\
        \midrule
        \circled{18} 
            & \makecell[l]{Perturbation Score\\Sensitivity (PertSS)}
            &  \multirow{1}{*}{VBCM} & \makecell[l]{$f(x,\; y(x)),\: A = \{x\}$}
            & \makecell{$| x - y |$} & $|T|$ & \makecell{$\{x_j\}$} \\
    \midrule 
        \circled{19}  
            & \makecell[l]{Perturbation Score\\Deviation (PertSD)}
            & \multirow{3}{*}{MCM}  & \makecell[l]{$f(x,\; y(x)),\: A = \{x\}$}
            &  \makecell{$\text{std}(X)$} & - & - \\
        \circled{20} 
            & \makecell[l]{Perturbation Score\\Range (PertSR)} 
            &  & \makecell[l]{$f(x,\; y(x)),\: A = \{x\}$} 
            & \makecell{$\text{max}(X) - \text{min}(X)$} & - & - \\
    \midrule 
        \circled{21} 
            & \makecell[l]{Average Individual\\Fairness (AvgIF)}
            & \multirow{2}{*}{PCM}  & \makecell[l]{\{$f(x, 1) \mid x \in A\}$} 
            & \makecell{$W_1(X, Y)$} & ${|T| \choose 2}$ &  - \\
        \ccircled{22} 
            & \makecell[l]{*Average Score\\Difference} 
            &  & \makecell[l]{mean($\{f(x, 1) \mid x \in A\}$)} 
            &  \makecell{$ x - y $}& ${|T| \choose 2}$ & - \\
    \bottomrule
    \end{tabular}
    
    \caption{Existing fairness metrics and how they fit in our generalized metrics. $f(x, c)$, $y(x)$ and $\hat{y}(x)$ are the probability associated with a class $c$, the gold class and the predicted class for example $x$, respectively.
    $MWU$ is the Mann-Whitney U test statistic and $W_1$ is the Wasserstein-1 distance between the distributions of $X$ and $Y$. Metrics marked with * have been defined in the context of only two protected groups and do not define the normalizing factor. {The metrics associated with gray circles cannot be applied to more than two groups (see \cref{sec:existing_metrics})}.
 \circled{1} \circled{2} \cite{dixon-2018}, \circled{3} \circled{21} \cite{huang-etal-2020-reducing}, \circled{4} \cite{beutel2019putting}, \circled{5} \circled{6} \cite{borkan_nuanced_2019}, \circled{7}  \cite{gaut-etal-2020-towards}, \circled{8}  \cite{Beutel2017DataDA, prost-etal-2019-debiasing}, \circled{9}  \cite{prost-etal-2019-debiasing}, \circled{10} \cite{Beutel2017DataDA},  \ccircled{11} \cite{Blodgett2017RacialDI, bhaskaran-bhallamudi-2019-good}, \ccircled{12} \cite{deartega2019},, \ccircled{13} \cite{stanovsky-etal-2019-evaluating, saunders-byrne-2020-reducing}, \ccircled{14} \cite{blodgett-etal-2018-twitter}, \ccircled{15} \cite{bamman-etal-2019-annotated}, \ccircled{16} \cite{webster-2018}, \circled{17}  \cite{garg-2019}, \circled{18}  \circled{19}  \circled{20}  \cite{prabhakaran-etal-2019-perturbation}, \ccircled{22} \cite{kiritchenko-mohammad-2018-examining, popovic2020joint}} 
    \label{tab:metrics-table}
    \vspace{1cm}
\end{table*}

\section{{Classifying Existing Fairness Metrics Within the Generalized Metrics}} \label{sec:existing_metrics}

\cref{tab:metrics-table} expresses 22 metrics from the literature as instances of our generalized metrics from \cref{gen-metrics}. {The presented metrics} span a number of NLP tasks, including text classification \cite{dixon-2018, kiritchenko-mohammad-2018-examining, garg-2019, borkan_nuanced_2019, prabhakaran-etal-2019-perturbation}, relation extraction \cite{gaut-etal-2020-towards}, text generation \cite{huang-etal-2020-reducing} and dependency parsing \cite{blodgett-etal-2018-twitter}.

We arrive at this list by reviewing 146 papers that study bias from the survey of \citet{blodgett_language_2020} and selecting metrics that meet three criteria: {(i) the metric is extrinsic; i.e., it is applied to at least one downstream NLP task,\footnote{{We do not consider language modeling to be a downstream task.}}} (ii) it quantifies the difference in performance across two or more groups, and (iii) it is not based on the \emphh{prediction} of a sensitive attribute{---metrics based on a model's predictions of sensitive attributes, e.g., in image captioning or text generation, constitute a specialized sub-type of fairness metrics.}
Out of the {26} metrics we find, only four do not fit within our framework: BPSN and BNSP \cite{borkan_nuanced_2019}, the {\small $\prod$} metric \cite{deartega2019} and Perturbation Label Distance \cite{prabhakaran-etal-2019-perturbation}.\footnote{BPSN and BNSP can be defined as Group VBCM if we relax the definition and allow for a separate $\phi$ function for the background---they require returning different confidence scores for the protected group and the background. The metrics of \newcite{prabhakaran-etal-2019-perturbation} \circled{18} \circled{19} \circled{21} originally have not been defined in terms of protected groups. In their paper, $T$ is a set of different names, both male and female.} 

Importantly, many of the metrics we find are PCMs defined for only two protected groups, typically for male and female genders or white and non-white races. Only those that use commutative $d$ can be straightforwardly adjusted to more groups. Those which cannot be adjusted are marked with gray circles in \cref{tab:metrics-table}.

\paragraph{Prediction v/s Probability Based Metrics} Beyond the categorization into PCM, BCM and MCM, as well as group and counterfactual fairness, the metrics can be further categorized into \emphh{prediction} or \emphh{probability} based. The former calculate the {score} based on a model's predictions, while the latter use the probabilities assigned to a particular class or label ({we found no metrics that make use of both probabilities and predictions}). 13 out of 16 group fairness metrics are prediction based, while \emphh{all} counterfactual metrics are probability based. Since the majority of metrics in \cref{tab:metrics-table} are defined for \emphh{binary} classification, the prevalent {scores} for prediction based metrics include false positive/negative rates (FPR/FNR) and true positive/negative rates (TPR/TNR).
Most of the probability-based metrics are based on the probability associated with the positive/toxic class (class 1 in binary classification). The exception are the metrics of \newcite{prabhakaran-etal-2019-perturbation} which utilize the probability of the \emphh{target} class \circled{18}  \circled{19}  \circled{21}.

\paragraph{Choice of $\phi$ and $d$} For scalar-valued $\phi$ the most common bi-variate comparison function is the (absolute) difference between two scores. As outliers, \newcite{beutel2019putting} \circled{4} use the ratio of the group score to the background score and \newcite{webster-2018} \ccircled{16} use the ratio between the first and the second group.
\newcite{prabhakaran-etal-2019-perturbation}'s MCM metrics use multivariate $d$. Their Perturbation Score Deviation metric \circled{19} uses the standard deviation of the scores, while their Perturbation Score Range metric \circled{20} uses the range of the scores (difference between the maximum and minimum score). For set-valued $\phi$, \newcite{huang-etal-2020-reducing} choose Wasserstein-1 distance \cite{jiang2020wasserstein} \circled{3} \circled{21}, while \newcite{borkan_nuanced_2019} define their comparison function using the Mann-Whitney U test statistic \cite{mann1947}.

\section{Experimental Details}
\label{sec:setup}

\begin{table}[t]
    \footnotesize
    \centering
    \renewcommand{\arraystretch}{1.2}

    \begin{tabular}{l| p{0.35\textwidth}}
    \toprule
    \makecell[l]{Sensitive\\attribute} & Protected groups ($T$) \\
    \midrule
            Gender & aab, female, male, cis, many-genders, no-gender, non-binary, trans\\
            \makecell[l]{Sexual\\Orientation} &
                asexual, 
                {homosexual},
                {heterosexual}, 
                {bisexual}, 
                {other} \\
            Religion &
                {atheism}, 
                {buddhism}, 
                {baha'i-faith}, 
                {christianity}, 
                {hinduism}, 
                {islam}, 
                {judaism}, 
                {mormonism}, 
                {sikhism}, 
                {taoism} \\
            Race &
                {african american}, 
                {american indian}, 
                {asian}, 
                {hispanic}, 
                {pacific islander}, 
                {white} \\
            Age & 
                young, 
                adult, 
                old \\
            Disability &
                {cerebral palsy}, 
                {chronic illness}, 
                {cognitive}, 
                {down syndrome}, 
                {epilepsy}, 
                {hearing}, 
                {mental health}, 
                {mobility}, 
                {physical}, 
                {short stature}, 
                {sight}, 
                {unspecified}, 
                {without}\\
            \midrule
            Nationality & We define 6 groups by categorizing countries based on their GDP. \\
    \bottomrule
    \end{tabular}
    \caption{The list of sensitive attributes and protected groups used in our experiments.}
    \label{tab:groups}
\end{table}

\begin{table}[t]
    \footnotesize
    \centering
    \renewcommand{\arraystretch}{1.2}

    \begin{tabular}{l | p{0.35\textwidth}}
    \toprule
    \makecell[l]{Protected\\group} & Identity terms ($I$)\\
    \midrule
        aab & AMAB, AFAB, DFAB, DMAB, female-assigned, male-assigned\\
        female & female (adj), female (n), woman\\
        male & male (adj), male (n), man\\
        \makecell[l]{many\\genders}& ambigender, ambigendered, androgynous, bigender, bigendered, intersex, intersexual, pangender, pangendered, polygender, androgyne, hermaphrodite\\
        no-gender & agender, agendered, genderless\\
    \bottomrule
    \end{tabular}
    \caption{Examples of explicit identity terms for the  selected protected groups of \word{Gender}.}
    \label{tab:terms}
\end{table}

{Having introduced our generalized framework and classified the existing metrics, we now \emphh{empirically} investigate their role in detecting the systemic performance difference across the demographic groups. We first discuss the relevant experimental details before presenting our results and analyses (\cref{sec:empirical_comp}).}

\begin{table}[t]
    \small
    \centering
    \begin{tabular}{c c | c}
    \toprule
        \makecell{SemEval-2} & \makecell{SemEval-3} & \makecell{CoNLL 2003}  \\
        \midrule
        \multicolumn{2}{c | }{Accuracy} & F1 \\
        \midrule
         0.90 & 0.73 & 0.94\\
    \bottomrule
    \end{tabular}
    \caption{RoBERTA performance on the official development splits for the three tasks.}
    \label{tab:models_res}
\end{table}

\paragraph{Models} We experiment on three RoBERTa~\cite{liu2019roberta} based models:\footnote{{Our preliminary experiments also used models based on Electra \cite{clark2020electra} as well as those trained on SST-2 and SST-3 datasets \cite{socher-etal-2013-recursive}. For all models, we observed similar trends in differences between the metrics. Due to space constraints we omit those results and leave a detailed cross-model bias comparison for future work.}}
(i) a binary classifier trained on SemEval-2018 valence classification shared task data \cite{SemEval2018Task1} processed for binary classification (SemEval-2)\footnote{We process the SemEval data as is commonly done for SST \cite{socher-etal-2013-recursive}. For binary classification, we filter out the neutral class and compress the multiple fine-grained positive/negative classes into a single positive/negative class. For 3-class classification we do not filter out the neutral class.} 
(ii) a 3-class classifier trained on SemEval-3 and (iii) a named entity recognition (NER) model trained on the CoNLL 2003 Shared Task data \cite{tjong-kim-sang-de-meulder-2003-introduction} which uses RoBERTa to encode a text sequence and a Conditional Random Field \cite{lafferty2001conditional} to predict the tags. {In NER experiments we use the \textsc{bilou} labeling scheme \cite{ratinov-roth-2009-design} and, for the probability-based metrics, we use the probabilities from the encoder's output.} Table \ref{tab:models_res} reports the performance on the {official} dev splits for the datasets the models were trained on.

\begin{table}[!t]
\setlength\tabcolsep{3pt}
\renewcommand{\arraystretch}{0.95}
\footnotesize
    \centering
    \begin{tabular}{ p{0.2cm} p{7cm} }
\toprule
         \multicolumn{2}{c }{Generic templates}\\
\midrule
 \rowcolor{palegreen} & I liked that \{identity\_adj\} writer. \\
\rowcolor{palegreen}   & As \{a:identity\_np\}, I feel hopeful. \\
 \rowcolor{palegreen} \multirow{-2}{*}{\rotatebox[origin=c]{90}{\textsc{pos}}}  & This summit is a wonderful event to celebrate \{identity\_adj\} folk. \\
 \multirow{3}{*}{\rotatebox[origin=c]{90}{\textsc{neut}}} & I recently found out that Alex is \{a:identity\_np\}. \\
   & The main character was \{identity\_adj\}. \\
   & I watched a documentary about \{identity\_adj\} writers. \\
 \rowcolor{palered}  & What a sad day to be \{identity\_adj\}. \\
  \rowcolor{palered} & It's scary how they treat \{identity\_adj\} people. \\
  \rowcolor{palered}  \multirow{-3}{*}{\rotatebox[origin=c]{90}{\textsc{neg}}} & Being \{identity\_adj\} makes life really hard. \\
\midrule
     \multicolumn{2}{c }{Nationality templates}\\
\midrule
\rowcolor{palegreen} & The next ACL will be in the beautiful \{country\}! \\
 \rowcolor{palegreen}  & The outstanding festival celebrated the best artists from \{country\}. \\
\rowcolor{palegreen}  \multirow{-3}{*}{\rotatebox[origin=c]{90}{\textsc{pos}}}  & \{Country\} has done a great job in fighting economic inequality. \\
  \multirow{3}{*}{\rotatebox[origin=c]{90}{\textsc{neut}}} & Alex always wanted to live in \{country\}.  \\
  & Next speaker travelled all the way from \{country\}. \\
  & \{Country\} is UK's most important trading partner. \\
  \rowcolor{palered}  & \{Country\} is the worst holiday destination. \\
  \rowcolor{palered} & \{Country\_adj\} Music Awards were a disaster.  \\
  \rowcolor{palered} \multirow{-2}{*}{\rotatebox[origin=c]{90}{\textsc{neg}}} & The designs of this \{country\_adj\} based firm are distressing. \\
\midrule
         \multicolumn{2}{c }{Name templates}\\
\midrule
\rowcolor{palegreen} & {Person} teared up and it was heartwarming. \\
 \rowcolor{palegreen}  & \{Person\} feels fulfilled as a parent.\\
 \rowcolor{palegreen} \multirow{-2}{*}{\rotatebox[origin=c]{90}{\textsc{pos}}}   & You should read this ground-breaking paper by \{person\}.\\
 \multirow{4}{*}{\rotatebox[origin=c]{90}{\textsc{neut}}} & As a child, \{person\} had big dreams.\\
  & \{Person\} received their high-school diploma in 1990. \\
  & \{Person\} came out with new strategies for managing and financing that program. \\
 \rowcolor{palered} & \{Person\} sadly didn't get the job. \\
 \rowcolor{palered}  & Nothing special, \{person\} thought so too. \\
 \rowcolor{palered}  \multirow{-2}{*}{\rotatebox[origin=c]{90}{\textsc{neg}}}  & It's unfair that \{person\} got that reputable position without prior experience. \\
\bottomrule
    \end{tabular}
    \caption{Template examples.}
    \label{tab:templates} \vspace{-0.1cm}
\end{table}

\paragraph{Evaluation Data} For classification, we experiment on seven sensitive attributes, and for each attribute we devise a number of protected groups (\cref{tab:groups}).\footnote{For Disability and Race we used the groups from \citet{hutchinson-etal-2020-social} and from the Racial and Ethnic Categories and Definitions for NIH Diversity Programs (\url{https://grants.nih.gov/grants/guide/notice-files/not-od-15-089.html}), respectively. For the remaining attributes, we rely on Wikipedia and Wiktionary, among other sources.} {We analyze bias within each attribute \emphh{independently} and focus on \emphh{explicit} mentions of each identity. This is reflected in our choice of identity terms, which we have gathered from Wikipedia, Wiktionary as well as \cite{dixon-2018} and \cite{hutchinson-etal-2020-social} (see \cref{tab:terms} for an example).} 
Additionally, for the \word{Gender} attribute we also investigate implicit mentions---{female} and {male} groups represented with names typically associated with these genders. 
We experiment on synthetic data created using hand-crafted templates, as is common in the literature \cite{dixon-2018, kiritchenko-mohammad-2018-examining, kurita-etal-2019-measuring, huang-etal-2020-reducing}.
For each sensitive attribute we use 60 templates with balanced classes; 20 negative, 20 neutral and 20 positive templates. For each attribute we use {30} generic templates---with adjective and noun phrase slots to be filled with identity terms---and {30} attribute-specific templates.\footnote{We will release all templates upon acceptance.} { In \cref{tab:templates} we present examples of both generic templates and attribute-specific templates for \word{Nationality}. Note that the slots of generic templates are designed to be filled with terms that explicitly reference an identity (\cref{tab:terms}), and are unsuitable for experiments on female/male names. For this reason, for names we design additional 30 name-specific templates (60 in total). We present examples of those templates in \cref{tab:templates}.}

{For NER, we only experiment on \word{Nationality} and generate the evaluation data from 22 templates with a missing \textit{\{country\}} slot for which we manually assign a \textsc{bilou} tag to each token. The \textit{\{country\}} slot is initially labeled as \textsc{u-loc} and is later automatically adjusted to a \emphh{sequence} of labels if a country name filling the slot spans more than one token, e.g., \textsc{b-loc} \textsc{l-loc} for \word{New Zeland}.}

\vspace{-0.05cm}
\paragraph{Metrics}
We experiment on metrics which support more than two protected groups (i.e., the \emphh{white-circled} metrics in \cref{tab:metrics-table}). {As described in \cref{sec:definitions}, for each source example we allow for a number of variations for each group. Hence, for counterfactual metrics which require only one example per group (all counterfactual metrics but Average Individual Fairness \circled{21}) we evaluate on the $|T|$-ary Cartesian products over the sets of variations for all groups. {For groups with large $|I|$ we sample 100 elements from the Cartesian product, without replacement.} We convert Counterfactual Token Fairness Gap \circled{17} and Perturbation Score Sensitivity \circled{18} 
into PCMs since for templated-data there is no single \emphh{real-world} example.
}

\begin{figure*}[!t]
  \centering
    \hspace{-0.3cm}
  \begin{subfigure}[b]{0.48\textwidth}
      \includegraphics[trim={0 13cm 0cm 0.1cm},clip,width=\textwidth]{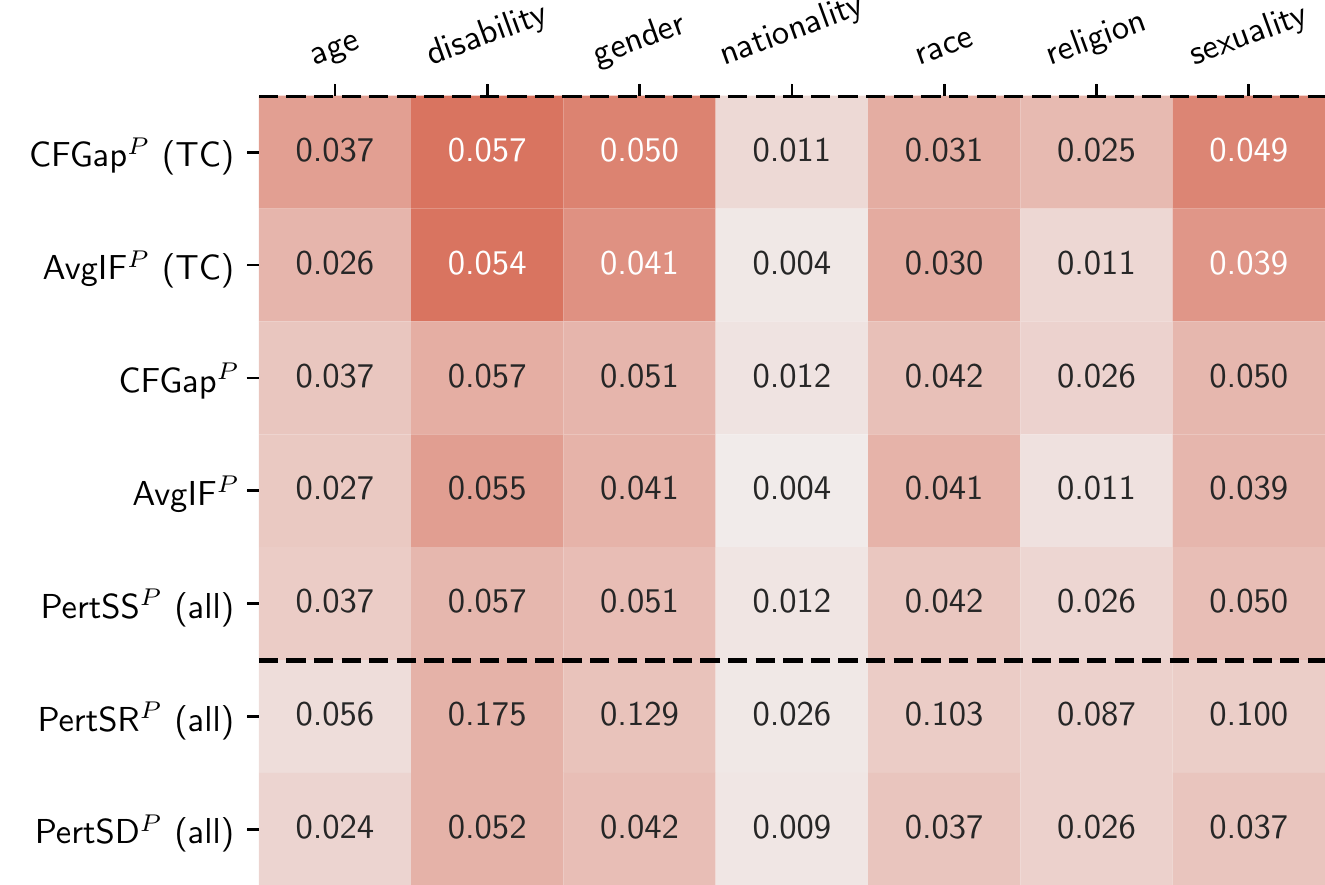}
      \caption{Counterfactual Metrics: SemEval-2}
  \end{subfigure}
  \begin{subfigure}[b]{0.388\textwidth}
  \hspace{0.3cm}
    \includegraphics[trim={2.55cm 13cm 0 0.1cm},clip,width=\textwidth]{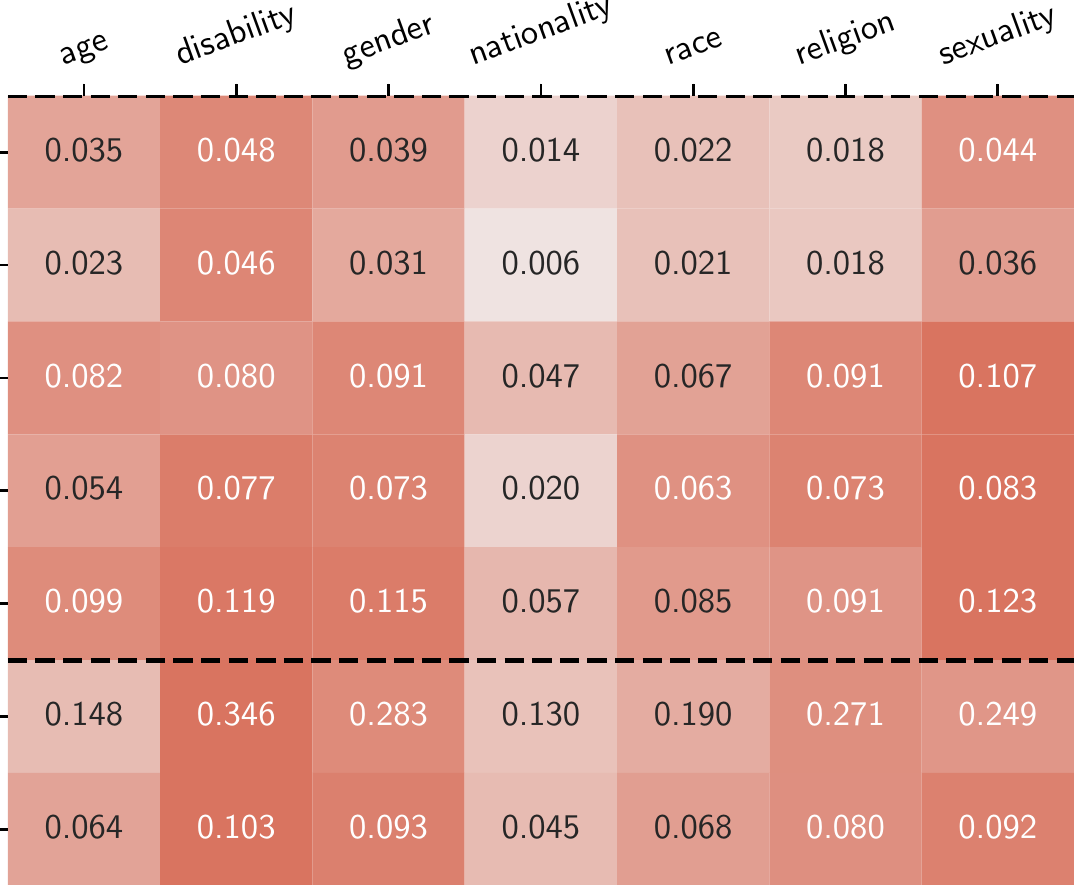}
    \caption{Counterfactual Metrics: SemEval-3}
  \end{subfigure}
  
  \vspace{0.1cm}
  
          \hspace{-0.3cm}
    \begin{subfigure}[b]{0.48\textwidth}
      \includegraphics[trim={0 0.05cm 0cm 9.5cm},clip,width=\textwidth]{all-attrs-semeval-2-scaled-with-semeval-3.pdf}
      \caption{Group Metrics: SemEval-2}
  \end{subfigure}
  \begin{subfigure}[b]{0.388\textwidth}
  \hspace{0.3cm}
    \includegraphics[trim={2.55cm 0.05cm 0 9.5cm},clip,width=\textwidth]{all-attrs-semeval-3-scaled-with-semeval-2.pdf}
    \caption{Group Metrics: SemEval-3}
  \end{subfigure}
  \vspace{-0.1cm}
  \caption{BCM, PCM and MCM metrics calculated for different sensitive attributes, for the positive class. Metrics marked with (all) are inherently multi-class and are calculated for all-classes.
  Superscripts {$^P$ and * mark the probability-based and correctly normalized metrics, respectively. We row-normalize the heatmap coloring, across the whole figure, using maximum absolute value scaling.}
  }
  \vspace{-0.45cm}
  \label{fig:all_attrs}
\end{figure*}

Average Group Fairness \circled{3}, Counterfacutal Token Fairnes Gap \circled{17} and Average Individual Fairness \circled{21}  calculate bias based on the probability of positive/toxic class on \emphh{all} examples. We introduce alternative versions of these metrics which calculate bias \emphh{only} on examples with gold label $c$, which we mark with a \word{(TC)} (for true class) suffix. 
The original versions target \word{demographic parity} \cite{dwork-2012}, while the TC versions target \word{equality of opportunity} \cite{hardt-2016} and can pinpoint the existence of bias more precisely, as we show later (\cref{sec:empirical_comp}).

\begin{figure*}[!t]
  \centering
\vspace{-0.1cm}
  \begin{subfigure}[b]{0.65\textwidth}
    \includegraphics[trim={0 0 0 0},clip,width=\textwidth]{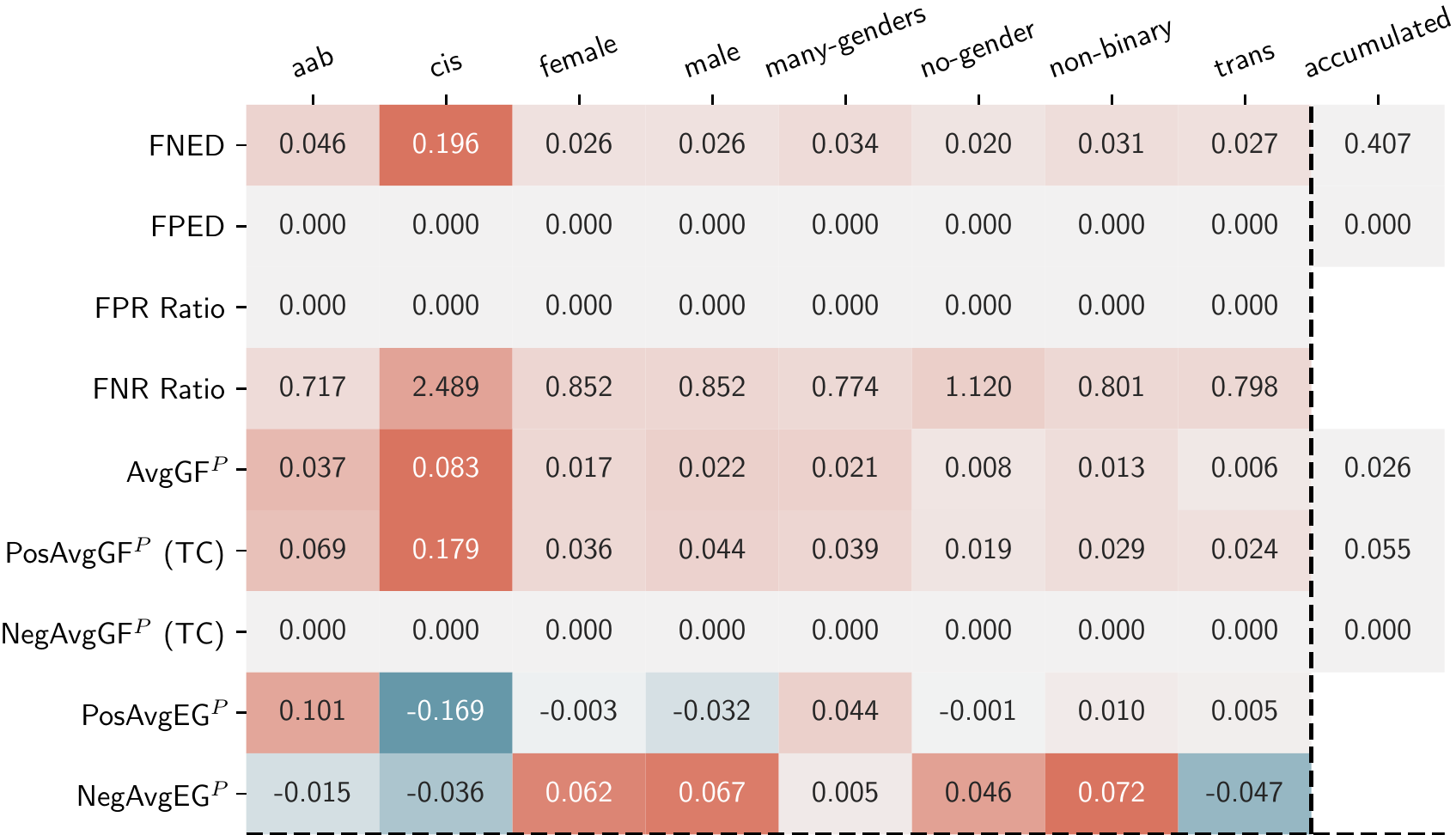}
    \caption{SemEval-2 (explicit identities)} \label{fig:gender_s2}
  \end{subfigure}
  \begin{subfigure}[b]{0.273\textwidth}
    \includegraphics[trim={2.7cm 0 0 0},clip,width=0.7\textwidth]{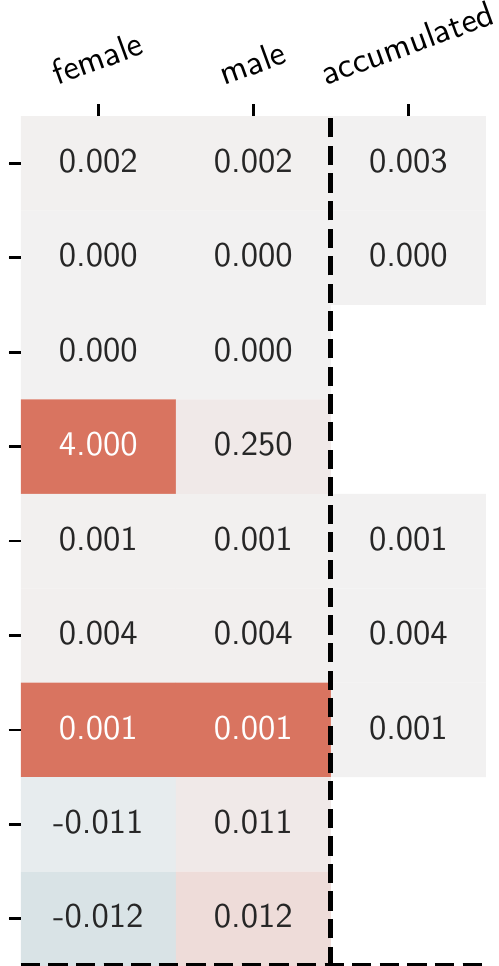}
    \caption{SemEval-2 (names)} \label{fig:gender_s2_names}
  \end{subfigure}
  \hspace{0.4cm}
  \begin{subfigure}[b]{0.675\textwidth}
    \includegraphics[trim={0 0 0 0},clip,width=0.95\textwidth]{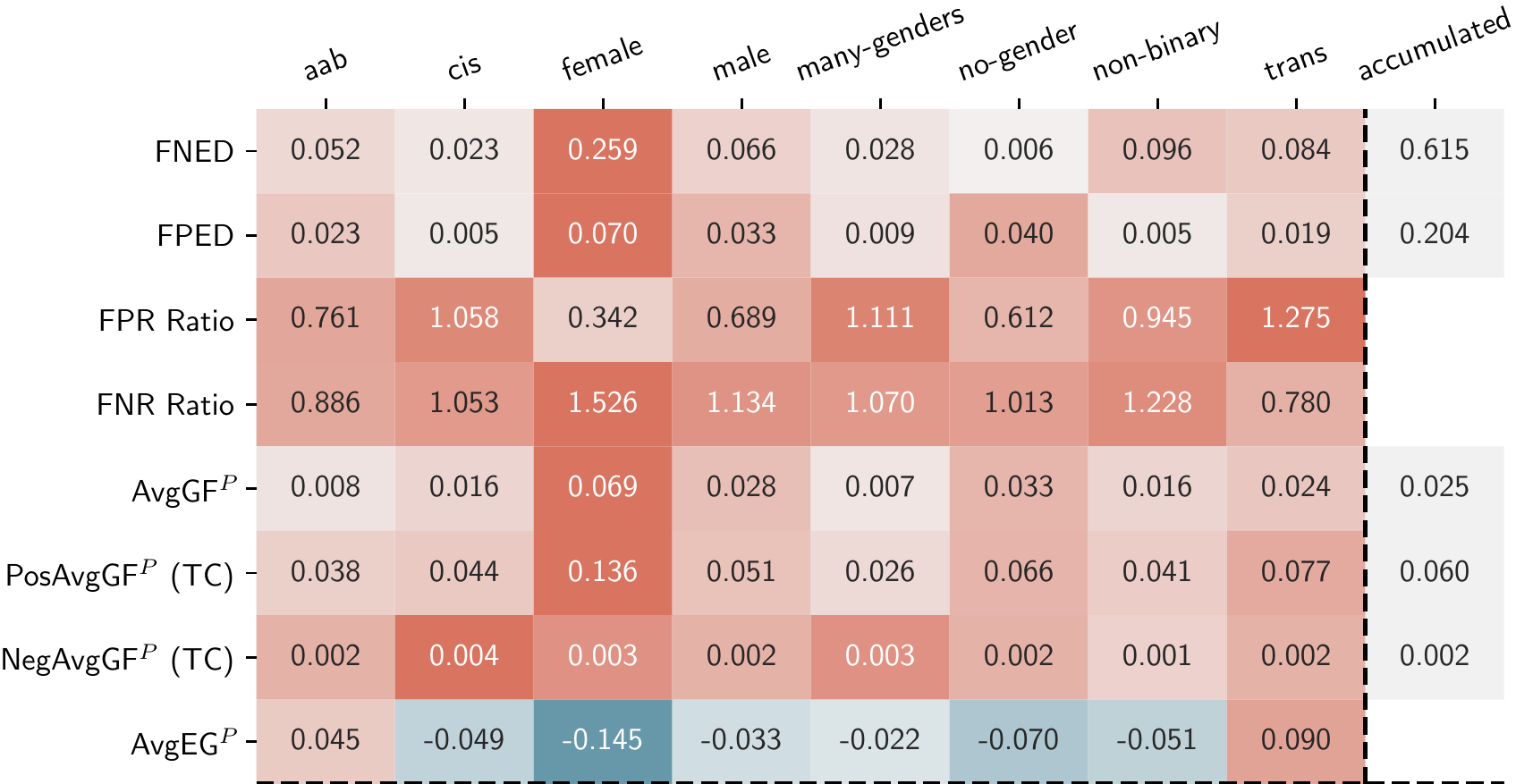}
    \caption{SemEval-3 (explicit identities)} \label{fig:gender_s3}
  \end{subfigure}
  \hspace{-0.44cm}
  \begin{subfigure}[b]{0.263\textwidth}
    \includegraphics[trim={2.8cm 0 0 0},clip,width=0.7\textwidth]{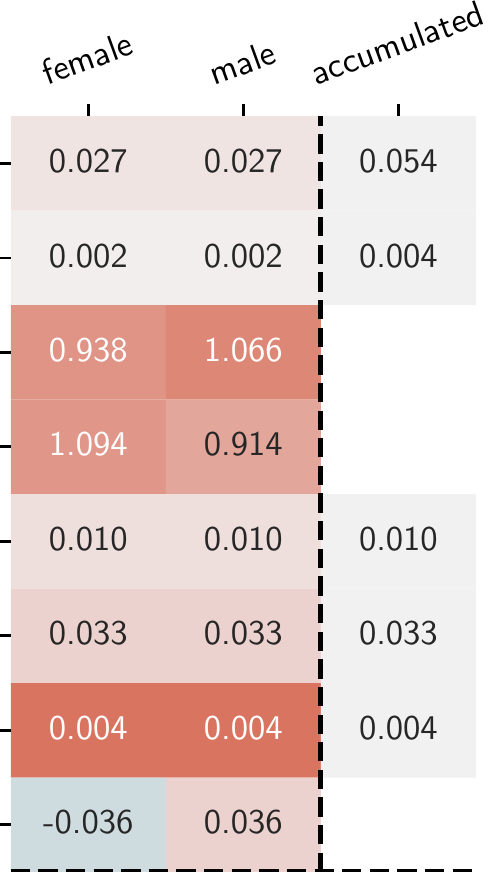}
    \caption{SemEval-3 (names) }\label{fig:gender_s3_names}
  \end{subfigure}
    \caption{Results for BCM and VBCM metrics on the positive class on \word{Gender} for explicit (left) and implicit identities, signaled through names (right).}  \label{fig:gender_semeval}
      \vspace{-0.45cm}
\end{figure*}

\subsection{Moving Beyond Binary Classification}
14 out of 15 \word{white-circled} metrics from \cref{tab:metrics-table} are inherently classification metrics, 11 of which are defined exclusively for binary classification. We adapt binary classification metrics to (i) multi-class classification and (ii) sequence labeling to support a broader range of NLP tasks.

\paragraph{Multi-class Classification}
Probability-based metrics that use the probability of the target class (\circled{18} \circled{19} \circled{20}) do not require any adaptations for multi-class classification. For other metrics, we measure bias independently for each class $c$, using a one-vs-rest strategy for prediction-based metrics and the probability of class $c$ for the scores of probability-based metrics  (\circled{3} \circled{5} \circled{6} \circled{17} \circled{21}).

\paragraph{Sequence Labeling}
 We view sequence labeling as a case of multi-class classification, with each token being a separate classification decision. As for multi-class classification, we compute the bias measurements for each class independently. 
For prediction-based metrics, we use one-vs-rest strategy and base the F1 and FNR scores on exact span matching.\footnote{We do not compute FPR based metrics, since false positives are unlikely to occur for our synthetic data and are less meaningful if they occur.} 
For probability-based metrics, \emphh{for each token} we accumulate the probability scores for different labels of the same class. E.g., with the \textsc{bilou} labeling scheme, the probabilities for \textsc{b-per}, \textsc{i-per}, \textsc{l-per} and \textsc{u-per} are summed to obtain the probability for the class \textsc{per}. Further, for counterfactual metrics, to account for different identity terms yielding different number of tokens, we average the probability scores for all tokens of multi-token identity terms.

\vspace{-0.1cm}
\section{Empirical Metric Comparison} \label{sec:empirical_comp}
\vspace{-0.1cm}

Fig. \ref{fig:all_attrs} shows the results for sentiment analysis for all attributes on BCM, PCM and MCM metrics. 
{In each table we report the original bias measurements and row-normalize the \emphh{heatmap coloring} using maximum absolute value scaling to allow for some cross-metric comparison.\footnote{{Even after normalization, bias measurements across metrics are not \emphh{fully} comparable---different metrics employ different base measurements (e.g., TPR, TNR etc.) and hence measure different aspects of bias.}}}
Fig. \ref{fig:all_attrs} gives evidence of unintended bias for most of the attributes we consider, with \word{Disability} and \word{Nationality} being the most and least affected attributes, respectively. {We highlight that since we evaluate on simple synthetic data in which the expressed sentiment is evident, even small performance differences can be concerning.} Fig. \ref{fig:all_attrs} also gives an initial insight into how the bias measurements vary across the metrics. 

\begin{figure}[!t]
    \centering
    
    \hspace{-0.38cm}
    \includegraphics[trim={0 0 0 0},clip,width=0.495\textwidth]{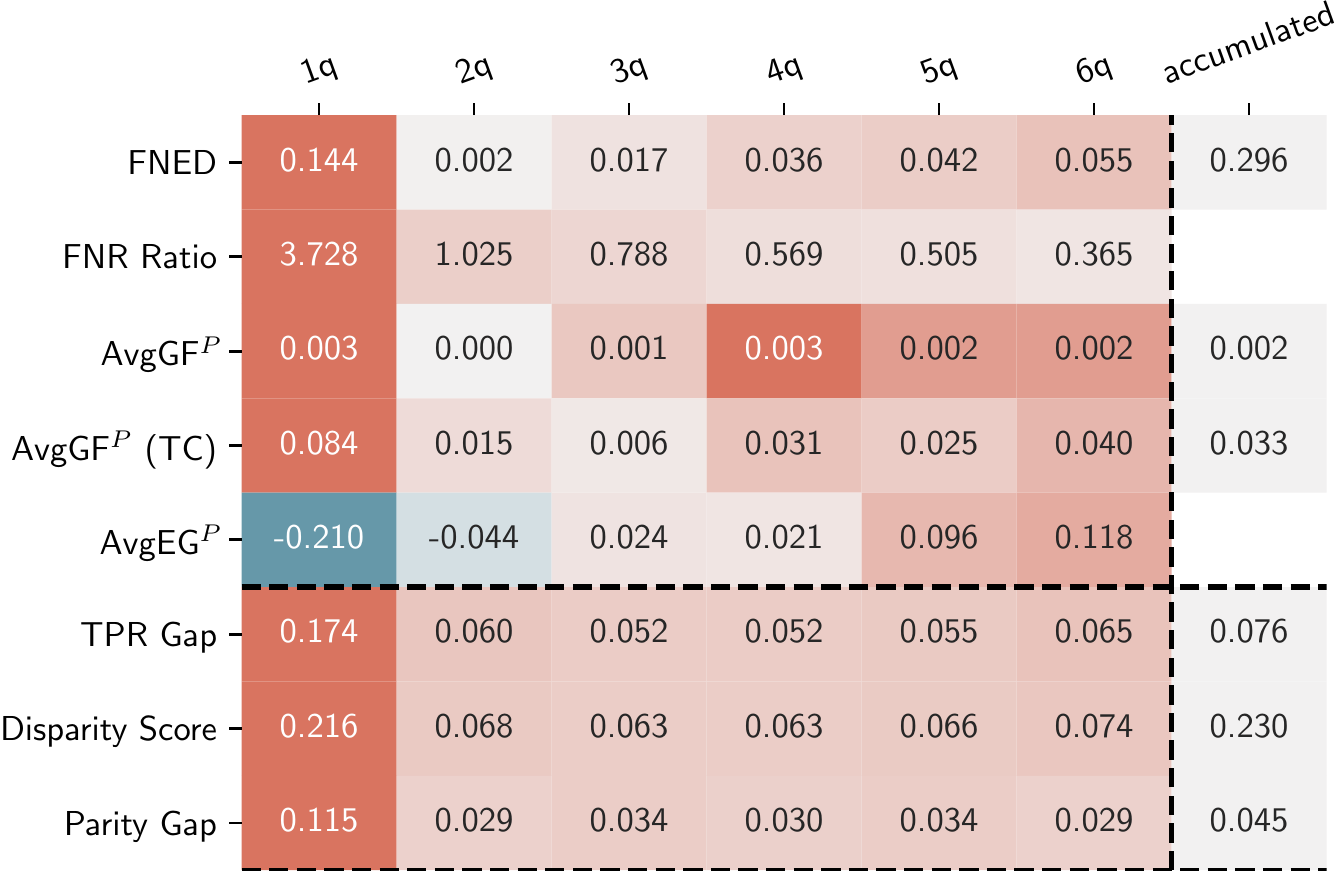}
    
    \vspace{0.1cm}
    
    \includegraphics[trim={0 0 0 0},clip,width=0.36\textwidth]{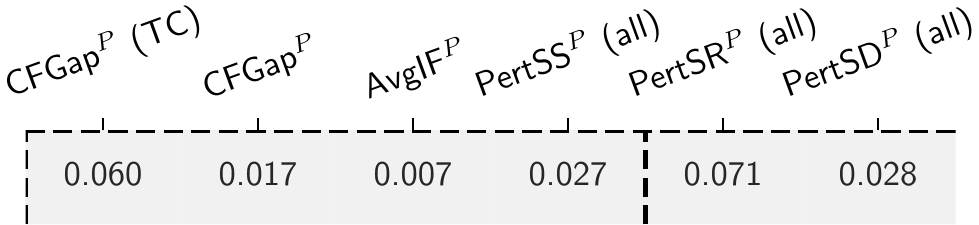}
    
    \caption{Results for the NER model on \word{Nationality} attribute for six groups defined by categorizing countries based on their GDP (six quantiles) for the (most relevant) \textsc{loc} class. We present group metrics at the top and the counterfactual metrics at the bottom. The probability-based metrics not marked with (TC) use probability scores for \textsc{loc} for \emphh{all} tokens, including \textsc{o}; hence they are less meaningful than their TC alternatives.}
    \label{fig:ner}
\end{figure}

In \cref{fig:gender_semeval} we present the per-group results for VBCM and BCM metrics for the {example} \word{Gender} attribute.\footnote{We omit the per-group results for the remaining attributes due to the lack of space. For BCM, we do not include \word{accumulated} values in the normalization.} Similarly, in \cref{fig:ner} we show results for NER for the relevant \textsc{loc} class. The first set of results indicates that the most problematic \word{Gender} group is \word{cis}. For NER we observe a big gap in the model's performance between the most affluent countries and countries with lower GDP. 
{In the context of those empirical results we now discuss how different parameter choices affect the observed bias measurement.}

\paragraph{Key Role of the Base Measurement}{ Perhaps the most important difference between the metrics lies in the parametrization of the scoring function $\phi$. The choice of $\phi$ determines what type and aspect of bias is being measured, making the metrics \emphh{conceptually} different.
Consider, for example $\phi$ of Average Group Fairness \circled{3}---$\{f(x, 1) \mid x \in A\}$---and Positive Average Equality Gap  \circled{5}---$\{f(x, 1) \mid x \in A, y(x) = 1\}$. They are both based on the probabilities associated with class 1, but the former is computed on \emphh{all} examples in $A$, while the latter is computed on only those examples that belong to the positive class (i.e. have gold label 1).
\ignore{but in contrast to the latter, the first is computed on \emphh{all} examples in $A$, not just those with gold class 1.} This difference causes them to measure {different types} of bias---the first targets \word{demographic parity}, the second \word{equality of opportunity}.

Further, consider FPED \circled{1} and FNED \circled{2} which employ FPR and FNR for their score, respectively. This difference alone can lead to entirely different results. E.g., in \cref{fig:gender_s2} FNED reveals prominent bias for the \word{cis} group while FPED shows none. Taken together, these results signal that the model's behaviour for this group \emphh{is} notably different from the other groups but this difference manifests itself \emphh{only} on the positive examples. 
}

\vspace{-0.1cm}
\paragraph{(In)Correct Normalization}
Next, we highlight the importance of correct normalization. We argue that fairness metrics should be invariant to the number of considered protected groups, otherwise the bias measurements are incomparable and can be misleadingly elevated. The latter is the case for three metrics---FPED \circled{1}, FNED \circled{2} and Disparity Score \circled{7}. The first two lack any kind of normalization, while Disparity Score is incorrectly normalized---$N$ is set to the number of groups, rather than group pairs. In \cref{fig:all_attrs} we present the results on the original versions of those metrics and for their correctly normalized versions, marked with *. The latter result in much lower bias measurements.
This is all the more important for FPED and FNED, as they have been very influential, with many works relying \emphh{exclusively} on these metrics \cite{rios2020fuzze, huang-etal-2020-multilingual,gencoglu2020cyberbullying, rios-lwowski-2020-empirical}.

\vspace{-0.05cm}
\paragraph{{Relative vs Absolute Comparison}}
Next, we argue that the results of metrics based on the relative comparison, e.g., FPR Ratio \circled{4}, can be misleading and hard to interpret if the original scores are not reported. In particular, the relative comparison can amplify bias in cases when both scores are low; in such scenario even a very small absolute difference can be relatively large. Such amplification is evident in the FNR Ratio metric (FNR equivalent of FPR Ratio) on {female} vs {male} names for RoBERTa fine-funed on SemEval-2 (\cref{fig:gender_s2_names}).  Similarly, when both scores are very high, the bias can be underestimated---a significant difference between the scores can seem relatively small if both scores are large. Indeed, such effects have also been widely discussed in the context of reporting health risks \cite{forrow1992absolutely, stegenga2015measuring, Noordzij2017RelativeRV}.
In contrast, the results of metrics based on absolute comparison can be meaningfully interpreted, even without the original scores, if the range of the scoring function is known and interpretable (which is the case for all metrics we review).

\vspace{-0.05cm}
\paragraph{Importance of Per-Group Results}
Most group metrics accumulate the results obtained for different groups. Such accumulation leads to diluted bias measurements in situations where the performance differs only for a small proportion of all groups. This is evident in, for example, the per-group NER results for correctly-normalized metrics (\cref{fig:ner}). 
We emphasize the importance of reporting per-group results whenever possible.

\vspace{-0.05cm}
\paragraph{Prediction vs Probability Based}
In contrast to prediction-based metrics, probability-based metrics capture {also more} subtle performance differences which do not lead to different predictions. This difference can be seen, for example, for \word{aab} \word{Gender} group results for SemEval-2 (\cref{fig:gender_s2}) and the results for {female}/{male} names for SemEval-3  (\cref{fig:gender_s3_names}). We contend it is beneficial to employ both types of metrics to understand the effect of behaviour differences {on predictions} and to allow for detection of more subtle differences. 

\vspace{-0.05cm}
\paragraph{Signed vs Unsigned}
Out of the 15 \emphh{white-circled} metrics only two are signed; Positive and Negative Average Equality Gap {(AvgEG)} \circled{5} \circled{6}.
Employing at least one signed metric allows for quick identification of the bias direction. For example, results for Average Equality Gap reveal that examples mentioning the \word{cis} \word{Gender} group are considered less positive than examples mentioning other groups and that, for NER, the probability of \textsc{loc} is \emphh{lower} for the richest countries (first and second quantiles have negative signs).

\paragraph{True Class Evaluation}
We observe that the TC versions of probability-metrics allow for better understanding of bias location, compared to their non-TC alternatives. Consider Average Group Fairness \circled{3} and its TC versions evaluated on the positive class (PosAvgGF) and negative class (NegAvgGF) for binary classification (\cref{fig:gender_s2}). The latter two reveal that the differences in behaviour apply solely to the positive examples.

\subsection{Fairness Metrics vs Significance Tests} \label{sec:significance}
Just like fairness metrics, statistical significance tests can also detect the presence of systematic differences in the behavior of a model, and hence are often employed as alternative means to quantify bias \cite{SemEval2018Task1, davidson-etal-2019-racial, Zhiltsova2019MitigationOU}. However, in contrast to fairness metrics, significance tests \emphh{do not} capture the magnitude of the differences. Rather, they quantify the likelihood of observing given differences under the null hypothesis. This is an important distinction with clear empirical consequences, as even very subtle differences between the scores can be statistically significant. 

To demonstrate this, we present p-values for significance tests for which we use the probability of the positive class as a dependent variable (\cref{tab:significance}). {Following \citet{kiritchenko-mohammad-2018-examining}, we obtain a single probability score for each template by averaging the results across all identity terms per group. Since we evaluate on synthetic data which is balanced across all groups, we use the scores for all templates regardless of their gold class.} We use the Friedman test for all attributes with more than two protected groups. For \word{Gender} with male/female names as identity terms we use the Wilcoxon signed-rank test.
We observe that, despite the low absolute values of the metrics obtained for the \word{Nationality} attribute (\cref{fig:all_attrs}), the behaviour of the models across the groups is unlikely to be equal. The same applies to the results for female vs male names for SemEval-3 (\cref{fig:gender_s3_names}). Employing a test for statistical significance can capture such nuanced presence of bias.

Notably, Average Equality Gap metrics \circled{5} \circled{6} occupy an atypical middle ground between being a fairness metric and a significance test. In contrast to other metrics from \cref{tab:metrics-table}, they \emphh{do not quantify the magnitude} of the differences, but the likelihood of a group being considered less positive than the background. 

\begin{table}[!t]
    \small
    \centering
    \begin{tabular}{l | l | l }
    \toprule
    Attribute & SemEval-2 & SemEval-3 \\
    \midrule
    Gender (names) & {$8.72 \times 10^{-1}$}  & $3.05 \times 10^{-6}$ \\
    Gender & $1.41 \times 10^{-8}$ & $3.80 \times 10^{-24}$ \\
    Sexual Orientation & $2.76 \times 10^{-9}$ &  $9.49 \times 10^{-24}$ \\
    Religion & $1.14 \times 10^{-23}$  & $8.24 \times 10^{-36}$ \\
    Nationality & $1.61 \times 10^{-2}$  & $1.45 \times 10^{-14}$  \\
    Race & $2.18 \times 10^{-5}$ &  $8.44 \times 10^{-5}$ \\
    Age & $4.86 \times 10^{-2}$ &  $4.81 \times 10^{-8}$  \\
    Disability & $9.67 \times 10^{-31}$ & $2.89 \times 10^{-44}$ \\
    \bottomrule
    \end{tabular}
    \caption{P-values for the Wilcoxon signed-rank test (attribute \word{Gender, (names)}) and the Friedman test (all other attributes).}
    \label{tab:significance}
\end{table}

\section{Which Metrics to Choose?} \label{sec:which}
In the previous section we highlighted important differences between the metrics which stem from different parameter choices. In particular, we emphasized the difference between prediction and probability-based metrics, in regards to their \word{sensitivity} to bias, as well as the conceptual distinction between the fairness metrics and significance tests. We also stressed the importance of correct normalization of metrics and reporting per-group results whenever possible. However, one important question still remains unanswered: out of the many different metrics that can be used, which ones are the most appropriate? Unfortunately, there is no easy answer. The choice of the metrics depends on many factors, including the task, the particulars of how and where the system is deployed, as well as the goals of the researcher. 

In line with the recommendations of \citet{olteanu2017limits} and \citet{blodgett_language_2020}, we assert that fairness metrics need to be grounded in the application domain and carefully matched to the type of studied bias to offer meaningful insights. 
While we cannot be too prescriptive about the exact metrics to choose, we advice against reporting results for all the metrics presented in this paper. Instead, we suggest a three-step process which helps to narrow down the full range of metrics to those that are the most applicable.

\paragraph{Step 1. Identifying the type of question to ask and choosing the appropriate generalized metric to answer it.} As discussed in \cref{gen-metrics}, each generalized metric is most suitable in different scenarios; e.g., MCM metrics can be used to investigate whether the attribute has any overall effect on the model's performance and (V)BCM allows to investigate how the performance for particular groups differs with respect to model's general performance.

\paragraph{Step 2. Identifying scoring functions which target the studied type and aspect of bias.}  At this stage it is important to consider practical consequences behind potential base measurements. E.g., for sentiment classification, misclassyfing positive sentences mentioning a specific demographic as negative can be more harmful than misclassyfing negative sentences as positive, as it can perpetuate negative stereotypes. Consequently, the most appropriate $\phi$ would be based on FNR or the probability of the negative class. In contrast, in the context of convicting low-level crimes, a false positive has more serious practical consequences than a false negative, since it may have a long-term detrimental effect on a person's life. Further, the parametrization of $\phi$ should be carefully matched to the motivation of the study and the assumed type/conceptualization of bias.

\paragraph{Step 3.} \textbf{Making the remaining parameter choices.} In particular, deciding on the comparison function most suitable for the selected $\phi$ and the targeted bias; e.g., absolute difference if $\phi$ is scalar-valued $\phi$ or Wasserstein-1 distance for set-valued $\phi$.
\vspace{0.2cm}

\noindent The above three steps can identify the most relevant metrics, which can be further filtered down to the minimal set sufficient to identify studied bias. To get a complete understanding of a model's (un)fairness, our general suggestion is to consider at least one prediction-based metric and one probability-based metric. Those can be further complemented with a test for statistical significance. Finally, it is essential that the results of each metric are interpreted in the context of the score employed by that metric (see \cref{sec:empirical_comp}). It is also universally good practice to report the results from all selected metrics, regardless of whether they do or do not give evidence of bias.

\section{Related Work}
{To our knowledge, we are the first to review and empirically compare fairness metrics used within NLP. Close to our endeavour are surveys which discuss types, sources and mitigation of bias in NLP or AI in general. Surveys of \citet{Mehrabi2019ASO},  \citet{hutchinson201950} and \citet{chouldechova2018frontiers} cover a broad scope of literature on algorithmic fairness. \citet{shah-etal-2020-predictive} offer both a survey of bias in NLP as well as a conceptual framework for studying bias.
\citet{sun2019mitigating} provide a comprehensive overview of addressing gender bias in NLP. There are also many task specific surveys, e.g., for language generation \cite{sheng2021societal} or machine translation \cite{Savoldi2021}. Finally, \citet{blodgett_language_2020} outline a number of methodological issues, such as providing vague motivations, which are common for papers on bias in NLP.

We focus on measuring bias exhibited on classification and sequence labeling downstream tasks. A related line of research measures bias present in sentence or word representations} \cite{bolukbasi-2016, caliskan-2017, kurita-etal-2019-measuring, sedoc-ungar-2019-role, chaloner-maldonado-2019-measuring, Dev2019AttenuatingBI, gonen-goldberg-2019-lipstick-pig,  hall-maudslay-etal-2019-name, liang-etal-2020-towards, shin-etal-2020-neutralizing, liang-etal-2020-towards, papakyriakopoulos-2020}. However, such intrinsic metrics have been recently shown not to correlate with application bias \cite{goldfarb2020intrinsic}. In yet another line of research, \newcite{badjatiya2019stereotypical} detect bias through identifying \defn{bias sensitive words}.

Beyond the fairness metrics and significance tests, some works quantify bias through calculating a standard evaluation metric, e.g., F1 or accuracy, or a more elaborate measure \emphh{independently} for each protected group or for each split of a challenge dataset \cite{hovy-sogaard-2015-tagging, rudinger-etal-2018-gender, zhao-etal-2018-gender, garimella-etal-2019-womens,  sap-etal-2019-risk,  bagdasaryan2019differential, stafanovics2020mitigating, tan-etal-2020-morphin, mehrabi-2020, Nadeem2020StereoSetMS, cao-daume-iii-2020-toward}.

\section{Conclusion}
We conduct a thorough review of existing fairness metrics and demonstrate that they are simply parametric variants of the {three generalized fairness metrics} we propose, {each suited to a different type of a scientific question. Further, we empirically demonstrate that the differences in parameter choices for our generalized metrics have direct impact on the bias measurement. In light of our results, we provide a range of concrete suggestions to guide NLP practitioners in their metric choices.

We hope that our work will facilitate further research in the bias domain and allow the researchers to direct their efforts towards bias mitigation.} Since our framework is language and model agnostic, in the future we plan to experiment on more languages and use our framework as principled means of comparing different models with respect to bias.

\section*{Acknowledgements}
We would like to thank the anonymous reviewers for their thoughtful comments and suggestions. We also thank the members of Amazon AI for many useful discussions and feedback.

\bibliography{tacl2018}

\begin{thebibliography}{79}
\expandafter\ifx\csname natexlab\endcsname\relax\def\natexlab#1{#1}\fi

\bibitem[{Badjatiya et~al.(2019)Badjatiya, Gupta, and
  Varma}]{badjatiya2019stereotypical}
Pinkesh Badjatiya, Manish Gupta, and Vasudeva Varma. 2019.
\newblock \href {https://dl.acm.org/doi/abs/10.1145/3308558.3313504}
  {{Stereotypical Bias Removal for Hate Speech Detection Task using
  Knowledge-based Generalizations}}.
\newblock In \emph{The World Wide Web Conference}, pages 49--59.

\bibitem[{Bagdasaryan et~al.(2019)Bagdasaryan, Poursaeed, and
  Shmatikov}]{bagdasaryan2019differential}
Eugene Bagdasaryan, Omid Poursaeed, and Vitaly Shmatikov. 2019.
\newblock \href
  {https://papers.nips.cc/paper/2019/file/fc0de4e0396fff257ea362983c2dda5a-Paper.pdf}
  {{Differential Privacy Has Disparate Impact on Model Accuracy}}.
\newblock In \emph{Advances in Neural Information Processing Systems}, pages
  15479--15488.

\bibitem[{Bamman et~al.(2019)Bamman, Popat, and
  Shen}]{bamman-etal-2019-annotated}
David Bamman, Sejal Popat, and Sheng Shen. 2019.
\newblock \href {https://doi.org/10.18653/v1/N19-1220} {An annotated dataset of
  literary entities}.
\newblock In \emph{Proceedings of the 2019 Conference of the North {A}merican
  Chapter of the Association for Computational Linguistics: Human Language
  Technologies, Volume 1 (Long and Short Papers)}, pages 2138--2144,
  Minneapolis, Minnesota. Association for Computational Linguistics.

\bibitem[{Beutel et~al.(2017)Beutel, Chen, Zhao, and hsin
  Chi}]{Beutel2017DataDA}
Alex Beutel, J.~Chen, Zhe Zhao, and Ed~Huai hsin Chi. 2017.
\newblock \href {https://arxiv.org/abs/1707.00075} {{Data Decisions and
  Theoretical Implications when Adversarially Learning Fair Representations}}.
\newblock \emph{Workshop on Fairness, Accountability, and Transparency in
  Machine Learning (FAT/ML 2017)}.

\bibitem[{Beutel et~al.(2019)Beutel, Chen, Doshi, Qian, Woodruff, Luu,
  Kreitmann, Bischof, and Chi}]{beutel2019putting}
Alex Beutel, Jilin Chen, Tulsee Doshi, Hai Qian, Allison Woodruff, Christine
  Luu, Pierre Kreitmann, Jonathan Bischof, and Ed~H Chi. 2019.
\newblock \href {https://dl.acm.org/doi/10.1145/3306618.3314234} {{Putting
  Fairness Principles into Practice: Challenges, Metrics, and Improvements}}.
\newblock In \emph{Proceedings of the 2019 AAAI/ACM Conference on AI, Ethics,
  and Society}, pages 453--459.

\bibitem[{Bhaskaran and Bhallamudi(2019)}]{bhaskaran-bhallamudi-2019-good}
Jayadev Bhaskaran and Isha Bhallamudi. 2019.
\newblock \href {https://doi.org/10.18653/v1/W19-3809} {{Good Secretaries, Bad
  Truck Drivers? Occupational Gender Stereotypes in Sentiment Analysis}}.
\newblock In \emph{Proceedings of the First Workshop on Gender Bias in Natural
  Language Processing}, pages 62--68, Florence, Italy. Association for
  Computational Linguistics.

\bibitem[{Blodgett et~al.(2020)Blodgett, Barocas, Daumé~III, and
  Wallach}]{blodgett_language_2020}
Su~Lin Blodgett, Solon Barocas, Hal Daumé~III, and Hanna Wallach. 2020.
\newblock \href {https://doi.org/10.18653/v1/2020.acl-main.485} {Language
  ({Technology}) is {Power}: {A} {Critical} {Survey} of "{Bias}" in {NLP}}.
\newblock In \emph{Proceedings of the 58th {Annual} {Meeting} of the
  {Association} for {Computational} {Linguistics}}, pages 5454--5476, Online.
  Association for Computational Linguistics.

\bibitem[{Blodgett and O'Connor(2017)}]{Blodgett2017RacialDI}
Su~Lin Blodgett and Brendan~T. O'Connor. 2017.
\newblock \href {https://arxiv.org/abs/1707.00061} {{Racial Disparity in
  Natural Language Processing: A Case Study of Social Media African-American
  English}}.
\newblock \emph{Workshop on Fairness, Accountability, and Transparency in
  Machine Learning (FAT/ML 2017)}.

\bibitem[{Blodgett et~al.(2018)Blodgett, Wei, and
  O{'}Connor}]{blodgett-etal-2018-twitter}
Su~Lin Blodgett, Johnny Wei, and Brendan O{'}Connor. 2018.
\newblock \href {https://doi.org/10.18653/v1/P18-1131} {{T}witter {U}niversal
  {D}ependency parsing for {A}frican-{A}merican and mainstream {A}merican
  {E}nglish}.
\newblock In \emph{Proceedings of the 56th Annual Meeting of the Association
  for Computational Linguistics (Volume 1: Long Papers)}, pages 1415--1425,
  Melbourne, Australia. Association for Computational Linguistics.

\bibitem[{Bolukbasi et~al.(2016)Bolukbasi, Chang, Zou, Saligrama, and
  Kalai}]{bolukbasi-2016}
Tolga Bolukbasi, Kai-Wei Chang, James Zou, Venkatesh Saligrama, and Adam Kalai.
  2016.
\newblock \href
  {https://papers.nips.cc/paper/2016/file/a486cd07e4ac3d270571622f4f316ec5-Paper.pdf}
  {{Man is to Computer Programmer as Woman is to Homemaker? Debiasing Word
  Embeddings}}.
\newblock In \emph{Proceedings of the 30th International Conference on Neural
  Information Processing Systems}, NIPS'16, pages 4356--4364, Red Hook, NY,
  USA. Curran Associates Inc.

\bibitem[{Borkan et~al.(2019)Borkan, Dixon, Sorensen, Thain, and
  Vasserman}]{borkan_nuanced_2019}
Daniel Borkan, Lucas Dixon, Jeffrey Sorensen, Nithum Thain, and Lucy Vasserman.
  2019.
\newblock \href {https://doi.org/10.1145/3308560.3317593} {Nuanced {Metrics}
  for {Measuring} {Unintended} {Bias} with {Real} {Data} for {Text}
  {Classification}}.
\newblock In \emph{Companion {Proceedings} of {The} 2019 {World} {Wide} {Web}
  {Conference}}, pages 491--500, San Francisco USA. ACM.

\bibitem[{Caliskan et~al.(2017)Caliskan, Bryson, and Narayanan}]{caliskan-2017}
Aylin Caliskan, Joanna Bryson, and Arvind Narayanan. 2017.
\newblock \href {https://science.sciencemag.org/content/356/6334/183}
  {Semantics derived automatically from language corpora contain human-like
  biases}.
\newblock \emph{Science}, 356:183--186.

\bibitem[{Cao and Daum{\'e}~III(2020)}]{cao-daume-iii-2020-toward}
Yang~Trista Cao and Hal Daum{\'e}~III. 2020.
\newblock \href {https://doi.org/10.18653/v1/2020.acl-main.418} {{Toward
  Gender-Inclusive Coreference Resolution}}.
\newblock In \emph{Proceedings of the 58th Annual Meeting of the Association
  for Computational Linguistics}, pages 4568--4595, Online. Association for
  Computational Linguistics.

\bibitem[{Chaloner and Maldonado(2019)}]{chaloner-maldonado-2019-measuring}
Kaytlin Chaloner and Alfredo Maldonado. 2019.
\newblock \href {https://doi.org/10.18653/v1/W19-3804} {{Measuring Gender Bias
  in Word Embeddings across Domains and Discovering New Gender Bias Word
  Categories}}.
\newblock In \emph{Proceedings of the First Workshop on Gender Bias in Natural
  Language Processing}, pages 25--32, Florence, Italy. Association for
  Computational Linguistics.

\bibitem[{Chouldechova and Roth(2020)}]{chouldechova2018frontiers}
Alexandra Chouldechova and Aaron Roth. 2020.
\newblock \href
  {https://cacm.acm.org/magazines/2020/5/244336-a-snapshot-of-the-frontiers-of-fairness-in-machine-learning/fulltext}
  {{A Snapshot of the Frontiers of Fairness in Machine Learning}}.
\newblock \emph{Communications of the ACM}, 63(5):82--89.

\bibitem[{Clark et~al.(2020)Clark, Luong, Le, and Manning}]{clark2020electra}
Kevin Clark, Minh-Thang Luong, Quoc~V Le, and Christopher~D Manning. 2020.
\newblock \href {https://openreview.net/pdf?id=r1xMH1BtvB} {{ELECTRA:
  Pre-training Text Encoders as Discriminators Rather Than Generators}}.
\newblock \emph{The International Conference on Learning Representations
  (ICLR)}.

\bibitem[{Davidson et~al.(2019)Davidson, Bhattacharya, and
  Weber}]{davidson-etal-2019-racial}
Thomas Davidson, Debasmita Bhattacharya, and Ingmar Weber. 2019.
\newblock \href {https://doi.org/10.18653/v1/W19-3504} {{Racial Bias in Hate
  Speech and Abusive Language Detection Datasets}}.
\newblock In \emph{Proceedings of the Third Workshop on Abusive Language
  Online}, pages 25--35, Florence, Italy. Association for Computational
  Linguistics.

\bibitem[{De-Arteaga et~al.(2019)De-Arteaga, Romanov, Wallach, Chayes, Borgs,
  Chouldechova, Geyik, Kenthapadi, and Kalai}]{deartega2019}
Maria De-Arteaga, Alexey Romanov, Hanna Wallach, Jennifer Chayes, Christian
  Borgs, Alexandra Chouldechova, Sahin Geyik, Krishnaram Kenthapadi, and
  Adam~Tauman Kalai. 2019.
\newblock \href {https://doi.org/10.1145/3287560.3287572} {{Bias in Bios: A
  Case Study of Semantic Representation Bias in a High-Stakes Setting}}.
\newblock In \emph{Proceedings of the Conference on Fairness, Accountability,
  and Transparency}, FAT* '19, pages 120--128, New York, NY, USA. Association
  for Computing Machinery.

\bibitem[{Dev and Phillips(2019)}]{Dev2019AttenuatingBI}
Sunipa Dev and Jeff Phillips. 2019.
\newblock \href {http://proceedings.mlr.press/v89/dev19a.html} {Attenuating
  bias in word vectors}.
\newblock In \emph{Proceedings of the Twenty-Second International Conference on
  Artificial Intelligence and Statistics}, volume~89 of \emph{Proceedings of
  Machine Learning Research}, pages 879--887. PMLR.

\bibitem[{Dixon et~al.(2018)Dixon, Li, Sorensen, Thain, and
  Vasserman}]{dixon-2018}
Lucas Dixon, John Li, Jeffrey Sorensen, Nithum Thain, and Lucy Vasserman. 2018.
\newblock \href {https://doi.org/10.1145/3278721.3278729} {{Measuring and
  Mitigating Unintended Bias in Text Classification}}.
\newblock In \emph{Proceedings of the 2018 AAAI/ACM Conference on AI, Ethics,
  and Society}, AIES '18, pages 67--73, New York, NY, USA. Association for
  Computing Machinery.

\bibitem[{Dwork et~al.(2012)Dwork, Hardt, Pitassi, Reingold, and
  Zemel}]{dwork-2012}
Cynthia Dwork, Moritz Hardt, Toniann Pitassi, Omer Reingold, and Richard Zemel.
  2012.
\newblock \href {https://doi.org/10.1145/2090236.2090255} {{Fairness through
  Awareness}}.
\newblock In \emph{Proceedings of the 3rd Innovations in Theoretical Computer
  Science Conference}, ITCS '12, pages 214--226, New York, NY, USA. Association
  for Computing Machinery.

\bibitem[{Forrow et~al.(1992)Forrow, Taylor, and Arnold}]{forrow1992absolutely}
Lachlan Forrow, William~C Taylor, and Robert~M Arnold. 1992.
\newblock \href {https://pubmed.ncbi.nlm.nih.gov/1543193/} {Absolutely
  relative: how research results are summarized can affect treatment
  decisions}.
\newblock \emph{The American journal of medicine}, 92(2):121--124.

\bibitem[{Garg et~al.(2019)Garg, Perot, Limtiaco, Taly, Chi, and
  Beutel}]{garg-2019}
Sahaj Garg, Vincent Perot, Nicole Limtiaco, Ankur Taly, Ed~H. Chi, and Alex
  Beutel. 2019.
\newblock \href {https://doi.org/10.1145/3306618.3317950} {{Counterfactual
  Fairness in Text Classification through Robustness}}.
\newblock In \emph{Proceedings of the 2019 AAAI/ACM Conference on AI, Ethics,
  and Society}, AIES '19, pages 219--226, New York, NY, USA. Association for
  Computing Machinery.

\bibitem[{Garimella et~al.(2019)Garimella, Banea, Hovy, and
  Mihalcea}]{garimella-etal-2019-womens}
Aparna Garimella, Carmen Banea, Dirk Hovy, and Rada Mihalcea. 2019.
\newblock \href {https://doi.org/10.18653/v1/P19-1339} {{Women{'}s Syntactic
  Resilience and Men{'}s Grammatical Luck: Gender-Bias in Part-of-Speech
  Tagging and Dependency Parsing}}.
\newblock In \emph{Proceedings of the 57th Annual Meeting of the Association
  for Computational Linguistics}, pages 3493--3498, Florence, Italy.
  Association for Computational Linguistics.

\bibitem[{Gaut et~al.(2020)Gaut, Sun, Tang, Huang, Qian, ElSherief, Zhao,
  Mirza, Belding, Chang, and Wang}]{gaut-etal-2020-towards}
Andrew Gaut, Tony Sun, Shirlyn Tang, Yuxin Huang, Jing Qian, Mai ElSherief,
  Jieyu Zhao, Diba Mirza, Elizabeth Belding, Kai-Wei Chang, and William~Yang
  Wang. 2020.
\newblock \href {https://doi.org/10.18653/v1/2020.acl-main.265} {{Towards
  Understanding Gender Bias in Relation Extraction}}.
\newblock In \emph{Proceedings of the 58th Annual Meeting of the Association
  for Computational Linguistics}, pages 2943--2953, Online. Association for
  Computational Linguistics.

\bibitem[{Gencoglu(2021)}]{gencoglu2020cyberbullying}
Oguzhan Gencoglu. 2021.
\newblock \href {https://doi.org/10.1109/MIC.2020.3032461} {{Cyberbullying
  Detection With Fairness Constraints}}.
\newblock \emph{IEEE Internet Computing}, 25(01):20--29.

\bibitem[{Goldfarb-Tarrant et~al.(2021)Goldfarb-Tarrant, Marchant, Sanchez,
  Pandya, and Lopez}]{goldfarb2020intrinsic}
Seraphina Goldfarb-Tarrant, Rebecca Marchant, Ricardo~Mu{\~n}oz Sanchez, Mugdha
  Pandya, and Adam Lopez. 2021.
\newblock \href {https://arxiv.org/abs/2012.15859} {{Intrinsic Bias Metrics Do
  Not Correlate with Application Bias}}.
\newblock In \emph{Proceedings of the 58th Annual Meeting of the Association
  for Computational Linguistics}. Association for Computational Linguistics.

\bibitem[{Gonen and Goldberg(2019)}]{gonen-goldberg-2019-lipstick-pig}
Hila Gonen and Yoav Goldberg. 2019.
\newblock \href {https://doi.org/10.18653/v1/N19-1061} {{Lipstick on a Pig:
  {D}ebiasing Methods Cover up Systematic Gender Biases in Word Embeddings But
  do not Remove Them}}.
\newblock In \emph{Proceedings of the 2019 Conference of the North {A}merican
  Chapter of the Association for Computational Linguistics: Human Language
  Technologies, Volume 1 (Long and Short Papers)}, pages 609--614, Minneapolis,
  Minnesota. Association for Computational Linguistics.

\bibitem[{Gonen and Webster(2020)}]{gonen-webster-2020-automatically}
Hila Gonen and Kellie Webster. 2020.
\newblock \href {https://doi.org/10.18653/v1/2020.findings-emnlp.180}
  {{Automatically Identifying Gender Issues in Machine Translation using
  Perturbations}}.
\newblock In \emph{Findings of the Association for Computational Linguistics:
  EMNLP 2020}, pages 1991--1995, Online. Association for Computational
  Linguistics.

\bibitem[{Hall~Maudslay et~al.(2019)Hall~Maudslay, Gonen, Cotterell, and
  Teufel}]{hall-maudslay-etal-2019-name}
Rowan Hall~Maudslay, Hila Gonen, Ryan Cotterell, and Simone Teufel. 2019.
\newblock \href {https://doi.org/10.18653/v1/D19-1530} {{It{'}s All in the
  Name: Mitigating Gender Bias with Name-Based Counterfactual Data
  Substitution}}.
\newblock In \emph{Proceedings of the 2019 Conference on Empirical Methods in
  Natural Language Processing and the 9th International Joint Conference on
  Natural Language Processing (EMNLP-IJCNLP)}, pages 5267--5275, Hong Kong,
  China. Association for Computational Linguistics.

\bibitem[{Hardt et~al.(2016)Hardt, Price, and Srebro}]{hardt-2016}
Moritz Hardt, Eric Price, and Nathan Srebro. 2016.
\newblock \href {https://dl.acm.org/doi/10.5555/3157382.3157469} {{Equality of
  Opportunity in Supervised Learning}}.
\newblock In \emph{Proceedings of the 30th International Conference on Neural
  Information Processing Systems}, NIPS'16, pages 3323--3331, Red Hook, NY,
  USA. Curran Associates Inc.

\bibitem[{Hovy and S{\o}gaard(2015)}]{hovy-sogaard-2015-tagging}
Dirk Hovy and Anders S{\o}gaard. 2015.
\newblock \href {https://doi.org/10.3115/v1/P15-2079} {{Tagging Performance
  Correlates with Author Age}}.
\newblock In \emph{Proceedings of the 53rd Annual Meeting of the Association
  for Computational Linguistics and the 7th International Joint Conference on
  Natural Language Processing (Volume 2: Short Papers)}, pages 483--488,
  Beijing, China. Association for Computational Linguistics.

\bibitem[{Huang et~al.(2020{\natexlab{a}})Huang, Zhang, Jiang, Stanforth,
  Welbl, Rae, Maini, Yogatama, and Kohli}]{huang-etal-2020-reducing}
Po-Sen Huang, Huan Zhang, Ray Jiang, Robert Stanforth, Johannes Welbl, Jack
  Rae, Vishal Maini, Dani Yogatama, and Pushmeet Kohli. 2020{\natexlab{a}}.
\newblock \href {https://doi.org/10.18653/v1/2020.findings-emnlp.7} {{Reducing
  Sentiment Bias in Language Models via Counterfactual Evaluation}}.
\newblock In \emph{Findings of the Association for Computational Linguistics:
  EMNLP 2020}, pages 65--83, Online. Association for Computational Linguistics.

\bibitem[{Huang et~al.(2020{\natexlab{b}})Huang, Xing, Dernoncourt, and
  Paul}]{huang-etal-2020-multilingual}
Xiaolei Huang, Linzi Xing, Franck Dernoncourt, and Michael~J. Paul.
  2020{\natexlab{b}}.
\newblock \href {https://www.aclweb.org/anthology/2020.lrec-1.180}
  {{Multilingual {T}witter Corpus and Baselines for Evaluating Demographic Bias
  in Hate Speech Recognition}}.
\newblock In \emph{Proceedings of the 12th Language Resources and Evaluation
  Conference}, pages 1440--1448, Marseille, France. European Language Resources
  Association.

\bibitem[{Hutchinson and Mitchell(2019)}]{hutchinson201950}
Ben Hutchinson and Margaret Mitchell. 2019.
\newblock \href {https://dl.acm.org/doi/10.1145/3287560.3287600} {{50 Years of
  Test (Un)fairness: Lessons for Machine Learning}}.
\newblock In \emph{Proceedings of the Conference on Fairness, Accountability,
  and Transparency}, pages 49--58.

\bibitem[{Hutchinson et~al.(2020)Hutchinson, Prabhakaran, Denton, Webster,
  Zhong, and Denuyl}]{hutchinson-etal-2020-social}
Ben Hutchinson, Vinodkumar Prabhakaran, Emily Denton, Kellie Webster, Yu~Zhong,
  and Stephen Denuyl. 2020.
\newblock \href {https://doi.org/10.18653/v1/2020.acl-main.487} {Social biases
  in {NLP} models as barriers for persons with disabilities}.
\newblock In \emph{Proceedings of the 58th Annual Meeting of the Association
  for Computational Linguistics}, pages 5491--5501, Online. Association for
  Computational Linguistics.

\bibitem[{Jacobs et~al.(2020)Jacobs, Blodgett, Barocas, Daum\'{e}, and
  Wallach}]{jacobs-2020}
Abigail~Z. Jacobs, Su~Lin Blodgett, Solon Barocas, Hal Daum\'{e}, and Hanna
  Wallach. 2020.
\newblock \href {https://doi.org/10.1145/3351095.3375671} {{The Meaning and
  Measurement of Bias: Lessons from Natural Language Processing}}.
\newblock In \emph{Proceedings of the 2020 Conference on Fairness,
  Accountability, and Transparency}, FAT* '20, page 706, New York, NY, USA.
  Association for Computing Machinery.

\bibitem[{Jiang et~al.(2020)Jiang, Pacchiano, Stepleton, Jiang, and
  Chiappa}]{jiang2020wasserstein}
Ray Jiang, Aldo Pacchiano, Tom Stepleton, Heinrich Jiang, and Silvia Chiappa.
  2020.
\newblock \href {http://proceedings.mlr.press/v115/jiang20a.html} {{Wasserstein
  Fair Classification}}.
\newblock In \emph{Proceedings of The 35th Uncertainty in Artificial
  Intelligence Conference}, volume 115 of \emph{Proceedings of Machine Learning
  Research}, pages 862--872. PMLR.

\bibitem[{{Kiritchenko} and
  {Mohammad}(2018)}]{kiritchenko-mohammad-2018-examining}
Svetlana {Kiritchenko} and Saif {Mohammad}. 2018.
\newblock \href {https://doi.org/10.18653/v1/S18-2005} {{Examining Gender and
  Race Bias in Two Hundred Sentiment Analysis Systems}}.
\newblock In \emph{Proceedings of the Seventh Joint Conference on Lexical and
  Computational Semantics}, pages 43--53, New Orleans, Louisiana. Association
  for Computational Linguistics.

\bibitem[{Kurita et~al.(2019)Kurita, Vyas, Pareek, Black, and
  Tsvetkov}]{kurita-etal-2019-measuring}
Keita Kurita, Nidhi Vyas, Ayush Pareek, Alan~W Black, and Yulia Tsvetkov. 2019.
\newblock \href {https://doi.org/10.18653/v1/W19-3823} {{Measuring Bias in
  Contextualized Word Representations}}.
\newblock In \emph{Proceedings of the First Workshop on Gender Bias in Natural
  Language Processing}, pages 166--172, Florence, Italy. Association for
  Computational Linguistics.

\bibitem[{Kusner et~al.(2017)Kusner, Loftus, Russell, and Silva}]{kusner-2017}
Matt~J Kusner, Joshua Loftus, Chris Russell, and Ricardo Silva. 2017.
\newblock \href
  {https://proceedings.neurips.cc/paper/2017/file/a486cd07e4ac3d270571622f4f316ec5-Paper.pdf}
  {{Counterfactual Fairness}}.
\newblock In \emph{Advances in Neural Information Processing Systems},
  volume~30, pages 4066--4076. Curran Associates, Inc.

\bibitem[{Lafferty et~al.(2001)Lafferty, McCallum, and
  Pereira}]{lafferty2001conditional}
John~D. Lafferty, Andrew McCallum, and Fernando C.~N. Pereira. 2001.
\newblock \href {https://dl.acm.org/doi/10.5555/645530.655813} {{Conditional
  Random Fields: Probabilistic Models for Segmenting and Labeling Sequence
  Data}}.
\newblock In \emph{Proceedings of the Eighteenth International Conference on
  Machine Learning}, ICML '01, page 282–289, San Francisco, CA, USA. Morgan
  Kaufmann Publishers Inc.

\bibitem[{Liang et~al.(2020)Liang, Li, Zheng, Lim, Salakhutdinov, and
  Morency}]{liang-etal-2020-towards}
Paul~Pu Liang, Irene~Mengze Li, Emily Zheng, Yao~Chong Lim, Ruslan
  Salakhutdinov, and Louis-Philippe Morency. 2020.
\newblock \href {https://doi.org/10.18653/v1/2020.acl-main.488} {Towards
  debiasing sentence representations}.
\newblock In \emph{Proceedings of the 58th Annual Meeting of the Association
  for Computational Linguistics}, pages 5502--5515, Online. Association for
  Computational Linguistics.

\bibitem[{Liu et~al.(2019)Liu, Ott, Goyal, Du, Joshi, Chen, Levy, Lewis,
  Zettlemoyer, and Stoyanov}]{liu2019roberta}
Yinhan Liu, Myle Ott, Naman Goyal, Jingfei Du, Mandar Joshi, Danqi Chen, Omer
  Levy, Mike Lewis, Luke Zettlemoyer, and Veselin Stoyanov. 2019.
\newblock \href {http://arxiv.org/abs/1907.11692} {{RoBERTa: A Robustly
  Optimized BERT Pretraining Approach}}.
\newblock \emph{CoRR}, abs/1907.11692. Version 1.

\bibitem[{Mann and Whitney(1947)}]{mann1947}
H.~B. Mann and D.~R. Whitney. 1947.
\newblock \href {https://doi.org/10.1214/aoms/1177730491} {{On a Test of
  Whether one of Two Random Variables is Stochastically Larger than the
  Other}}.
\newblock \emph{Ann. Math. Statist.}, 18(1):50--60.

\bibitem[{Mehrabi et~al.(2020)Mehrabi, Gowda, Morstatter, Peng, and
  Galstyan}]{mehrabi-2020}
Ninareh Mehrabi, Thamme Gowda, Fred Morstatter, Nanyun Peng, and Aram Galstyan.
  2020.
\newblock \href {https://doi.org/10.1145/3372923.3404804} {{Man is to Person as
  Woman is to Location: Measuring Gender Bias in Named Entity Recognition}}.
\newblock In \emph{Proceedings of the 31st ACM Conference on Hypertext and
  Social Media}, HT '20, pages 231--232, New York, NY, USA. Association for
  Computing Machinery.

\bibitem[{Mehrabi et~al.(2019)Mehrabi, Morstatter, Saxena, Lerman, and
  Galstyan}]{Mehrabi2019ASO}
Ninareh Mehrabi, Fred Morstatter, N.~Saxena, Kristina Lerman, and A.~Galstyan.
  2019.
\newblock \href {http://arxiv.org/abs/1908.09635} {{A Survey on Bias and
  Fairness in Machine Learning}}.
\newblock \emph{CoRR}, abs/1908.09635. Version 2.

\bibitem[{Mohammad et~al.(2018)Mohammad, Bravo-Marquez, Salameh, and
  Kiritchenko}]{SemEval2018Task1}
Saif Mohammad, Felipe Bravo-Marquez, Mohammad Salameh, and Svetlana
  Kiritchenko. 2018.
\newblock \href {https://doi.org/10.18653/v1/S18-1001} {{{S}em{E}val-2018 Task
  1: Affect in Tweets}}.
\newblock In \emph{Proceedings of The 12th International Workshop on Semantic
  Evaluation}, pages 1--17, New Orleans, Louisiana. Association for
  Computational Linguistics.

\bibitem[{Nadeem et~al.(2020)Nadeem, Bethke, and Reddy}]{Nadeem2020StereoSetMS}
Moin Nadeem, Anna Bethke, and Siva Reddy. 2020.
\newblock \href {https://arxiv.org/abs/2004.09456} {{StereoSet: Measuring
  stereotypical bias in pretrained language models}}.
\newblock \emph{CoRR}, abs/2004.09456. Version 1.

\bibitem[{Nangia et~al.(2020)Nangia, Vania, Bhalerao, and
  Bowman}]{nangia-etal-2020-crows}
Nikita Nangia, Clara Vania, Rasika Bhalerao, and Samuel~R. Bowman. 2020.
\newblock \href {https://doi.org/10.18653/v1/2020.emnlp-main.154}
  {{{C}row{S}-Pairs: A Challenge Dataset for Measuring Social Biases in Masked
  Language Models}}.
\newblock In \emph{Proceedings of the 2020 Conference on Empirical Methods in
  Natural Language Processing (EMNLP)}, pages 1953--1967, Online. Association
  for Computational Linguistics.

\bibitem[{Noordzij et~al.(2017)Noordzij, van Diepen, Caskey, and
  Jager}]{Noordzij2017RelativeRV}
M.~Noordzij, M.~van Diepen, F.~Caskey, and K.~Jager. 2017.
\newblock \href {https://pubmed.ncbi.nlm.nih.gov/28339913/} {Relative risk
  versus absolute risk: one cannot be interpreted without the other: Clinical
  epidemiology in nephrology}.
\newblock \emph{Nephrology Dialysis Transplantation}, 32:ii13–ii18.

\bibitem[{Olteanu et~al.(2017)Olteanu, Talamadupula, and
  Varshney}]{olteanu2017limits}
Alexandra Olteanu, Kartik Talamadupula, and Kush~R Varshney. 2017.
\newblock \href {https://dl.acm.org/doi/10.1145/3091478.3098871} {The limits of
  abstract evaluation metrics: The case of hate speech detection}.
\newblock In \emph{Proceedings of the 2017 ACM on Web Science Conference},
  pages 405--406.

\bibitem[{Papakyriakopoulos et~al.(2020)Papakyriakopoulos, Hegelich, Serrano,
  and Marco}]{papakyriakopoulos-2020}
Orestis Papakyriakopoulos, Simon Hegelich, Juan Carlos~Medina Serrano, and
  Fabienne Marco. 2020.
\newblock \href {https://doi.org/10.1145/3351095.3372843} {{Bias in Word
  Embeddings}}.
\newblock In \emph{Proceedings of the 2020 Conference on Fairness,
  Accountability, and Transparency}, FAT* '20, pages 446--457, New York, NY,
  USA. Association for Computing Machinery.

\bibitem[{Popovi{\'{c}} et~al.(2020)Popovi{\'{c}}, Lemmerich, and
  Strohmaier}]{popovic2020joint}
Radomir Popovi{\'{c}}, Florian Lemmerich, and Markus Strohmaier. 2020.
\newblock \href {https://link.springer.com/chapter/10.1007/978-3-030-59491-6_8}
  {{Joint Multiclass Debiasing of Word Embeddings}}.
\newblock In \emph{Foundations of Intelligent Systems}, pages 79--89, Cham.
  Springer International Publishing.

\bibitem[{Prabhakaran et~al.(2019)Prabhakaran, Hutchinson, and
  Mitchell}]{prabhakaran-etal-2019-perturbation}
Vinodkumar Prabhakaran, Ben Hutchinson, and Margaret Mitchell. 2019.
\newblock \href {https://doi.org/10.18653/v1/D19-1578} {{Perturbation
  Sensitivity Analysis to Detect Unintended Model Biases}}.
\newblock In \emph{Proceedings of the 2019 Conference on Empirical Methods in
  Natural Language Processing and the 9th International Joint Conference on
  Natural Language Processing (EMNLP-IJCNLP)}, pages 5740--5745, Hong Kong,
  China. Association for Computational Linguistics.

\bibitem[{Prost et~al.(2019)Prost, Thain, and
  Bolukbasi}]{prost-etal-2019-debiasing}
Flavien Prost, Nithum Thain, and Tolga Bolukbasi. 2019.
\newblock \href {https://doi.org/10.18653/v1/W19-3810} {{Debiasing Embeddings
  for Reduced Gender Bias in Text Classification}}.
\newblock In \emph{Proceedings of the First Workshop on Gender Bias in Natural
  Language Processing}, pages 69--75, Florence, Italy. Association for
  Computational Linguistics.

\bibitem[{Ratinov and Roth(2009)}]{ratinov-roth-2009-design}
Lev Ratinov and Dan Roth. 2009.
\newblock \href {https://www.aclweb.org/anthology/W09-1119} {{Design Challenges
  and Misconceptions in Named Entity Recognition}}.
\newblock In \emph{Proceedings of the Thirteenth Conference on Computational
  Natural Language Learning ({C}o{NLL}-2009)}, pages 147--155, Boulder,
  Colorado. Association for Computational Linguistics.

\bibitem[{Ribeiro et~al.(2020)Ribeiro, Wu, Guestrin, and
  Singh}]{ribeiro_beyond_2020}
Marco~Tulio Ribeiro, Tongshuang Wu, Carlos Guestrin, and Sameer Singh. 2020.
\newblock \href {https://doi.org/10.18653/v1/2020.acl-main.442} {Beyond
  {Accuracy}: {Behavioral} {Testing} of {NLP} {Models} with {CheckList}}.
\newblock In \emph{Proceedings of the 58th {Annual} {Meeting} of the
  {Association} for {Computational} {Linguistics}}, pages 4902--4912, Online.
  Association for Computational Linguistics.

\bibitem[{Rios(2020)}]{rios2020fuzze}
Anthony Rios. 2020.
\newblock \href {https://ojs.aaai.org//index.php/AAAI/article/view/5434}
  {{FuzzE: Fuzzy Fairness Evaluation of Offensive Language Classifiers on
  African-American English}}.
\newblock In \emph{Proceedings of the AAAI Conference on Artificial
  Intelligence}, volume~34, pages 881--889.

\bibitem[{Rios and Lwowski(2020)}]{rios-lwowski-2020-empirical}
Anthony Rios and Brandon Lwowski. 2020.
\newblock \href {https://www.aclweb.org/anthology/2020.coling-main.299} {{An
  Empirical Study of the Downstream Reliability of Pre-Trained Word
  Embeddings}}.
\newblock In \emph{Proceedings of the 28th International Conference on
  Computational Linguistics}, pages 3371--3388, Barcelona, Spain (Online).
  International Committee on Computational Linguistics.

\bibitem[{Rudinger et~al.(2018)Rudinger, Naradowsky, Leonard, and
  Van~Durme}]{rudinger-etal-2018-gender}
Rachel Rudinger, Jason Naradowsky, Brian Leonard, and Benjamin Van~Durme. 2018.
\newblock \href {https://doi.org/10.18653/v1/N18-2002} {Gender bias in
  coreference resolution}.
\newblock In \emph{Proceedings of the 2018 Conference of the North {A}merican
  Chapter of the Association for Computational Linguistics: Human Language
  Technologies, Volume 2 (Short Papers)}, pages 8--14, New Orleans, Louisiana.
  Association for Computational Linguistics.

\bibitem[{Sap et~al.(2019)Sap, Card, Gabriel, Choi, and
  Smith}]{sap-etal-2019-risk}
Maarten Sap, Dallas Card, Saadia Gabriel, Yejin Choi, and Noah~A. Smith. 2019.
\newblock \href {https://doi.org/10.18653/v1/P19-1163} {{The Risk of Racial
  Bias in Hate Speech Detection}}.
\newblock In \emph{Proceedings of the 57th Annual Meeting of the Association
  for Computational Linguistics}, pages 1668--1678, Florence, Italy.
  Association for Computational Linguistics.

\bibitem[{Saunders and Byrne(2020)}]{saunders-byrne-2020-reducing}
Danielle Saunders and Bill Byrne. 2020.
\newblock \href {https://doi.org/10.18653/v1/2020.acl-main.690} {{Reducing
  Gender Bias in Neural Machine Translation as a Domain Adaptation Problem}}.
\newblock In \emph{Proceedings of the 58th Annual Meeting of the Association
  for Computational Linguistics}, pages 7724--7736, Online. Association for
  Computational Linguistics.

\bibitem[{Savoldi et~al.(2021)Savoldi, Gaido, Bentivogli, Negri, and
  Turchi}]{Savoldi2021}
Beatrice Savoldi, Marco Gaido, Luisa Bentivogli, Matteo Negri, and Marco
  Turchi. 2021.
\newblock \href {https://arxiv.org/abs/2104.06001} {{Gender Bias in Machine
  Translation}}.
\newblock \emph{Transactions of the Association for Computational Linguistics}.

\bibitem[{Sedoc and Ungar(2019)}]{sedoc-ungar-2019-role}
Jo{\~a}o Sedoc and Lyle Ungar. 2019.
\newblock \href {https://doi.org/10.18653/v1/W19-3808} {{The Role of Protected
  Class Word Lists in Bias Identification of Contextualized Word
  Representations}}.
\newblock In \emph{Proceedings of the First Workshop on Gender Bias in Natural
  Language Processing}, pages 55--61, Florence, Italy. Association for
  Computational Linguistics.

\bibitem[{Shah et~al.(2020)Shah, Schwartz, and
  Hovy}]{shah-etal-2020-predictive}
Deven~Santosh Shah, H.~Andrew Schwartz, and Dirk Hovy. 2020.
\newblock \href {https://doi.org/10.18653/v1/2020.acl-main.468} {{Predictive
  Biases in Natural Language Processing Models: A Conceptual Framework and
  Overview}}.
\newblock In \emph{Proceedings of the 58th Annual Meeting of the Association
  for Computational Linguistics}, pages 5248--5264, Online. Association for
  Computational Linguistics.

\bibitem[{Sheng et~al.(2020)Sheng, Chang, Natarajan, and
  Peng}]{sheng-etal-2020-towards}
Emily Sheng, Kai-Wei Chang, Prem Natarajan, and Nanyun Peng. 2020.
\newblock \href {https://doi.org/10.18653/v1/2020.findings-emnlp.291} {Towards
  {C}ontrollable {B}iases in {L}anguage {G}eneration}.
\newblock In \emph{Findings of the Association for Computational Linguistics:
  EMNLP 2020}, pages 3239--3254, Online. Association for Computational
  Linguistics.

\bibitem[{Sheng et~al.(2021)Sheng, Chang, Natarajan, and
  Peng}]{sheng2021societal}
Emily Sheng, Kai-Wei Chang, Premkumar Natarajan, and Nanyun Peng. 2021.
\newblock \href {https://arxiv.org/abs/2105.04054} {{Societal Biases in
  Language Generation: Progress and Challenges}}.
\newblock In \emph{Proceedings of the 58th Annual Meeting of the Association
  for Computational Linguistics}. Association for Computational Linguistics.

\bibitem[{Shin et~al.(2020)Shin, Song, Jang, Kim, Joo, and
  Moon}]{shin-etal-2020-neutralizing}
Seungjae Shin, Kyungwoo Song, JoonHo Jang, Hyemi Kim, Weonyoung Joo, and
  Il-Chul Moon. 2020.
\newblock \href {https://doi.org/10.18653/v1/2020.findings-emnlp.280}
  {{Neutralizing Gender Bias in Word Embeddings with Latent Disentanglement and
  Counterfactual Generation}}.
\newblock In \emph{Findings of the Association for Computational Linguistics:
  EMNLP 2020}, pages 3126--3140, Online. Association for Computational
  Linguistics.

\bibitem[{Socher et~al.(2013)Socher, Perelygin, Wu, Chuang, Manning, Ng, and
  Potts}]{socher-etal-2013-recursive}
Richard Socher, Alex Perelygin, Jean Wu, Jason Chuang, Christopher~D. Manning,
  Andrew Ng, and Christopher Potts. 2013.
\newblock \href {https://www.aclweb.org/anthology/D13-1170} {{Recursive Deep
  Models for Semantic Compositionality Over a Sentiment Treebank}}.
\newblock In \emph{Proceedings of the 2013 Conference on Empirical Methods in
  Natural Language Processing}, pages 1631--1642, Seattle, Washington, USA.
  Association for Computational Linguistics.

\bibitem[{Stafanovi\v{c}s et~al.(2020)Stafanovi\v{c}s, Bergmanis, and
  Pinnis}]{stafanovics2020mitigating}
Art\={u}rs Stafanovi\v{c}s, Toms Bergmanis, and M\={a}rcis Pinnis. 2020.
\newblock \href {https://www.aclweb.org/anthology/P19-1163} {{Mitigating Gender
  Bias in Machine Translation with Target Gender Annotations}}.
\newblock In \emph{Proceedings of the 5th Conference on Machine Translation
  (WMT)}, pages 629--638. Association for Computational Linguistics.

\bibitem[{Stanovsky et~al.(2019)Stanovsky, Smith, and
  Zettlemoyer}]{stanovsky-etal-2019-evaluating}
Gabriel Stanovsky, Noah~A. Smith, and Luke Zettlemoyer. 2019.
\newblock \href {https://doi.org/10.18653/v1/P19-1164} {{Evaluating Gender Bias
  in Machine Translation}}.
\newblock In \emph{Proceedings of the 57th Annual Meeting of the Association
  for Computational Linguistics}, pages 1679--1684, Florence, Italy.
  Association for Computational Linguistics.

\bibitem[{Stegenga(2015)}]{stegenga2015measuring}
Jacob Stegenga. 2015.
\newblock \href
  {https://www.sciencedirect.com/science/article/abs/pii/S1369848615000837}
  {Measuring effectiveness}.
\newblock \emph{Studies in History and Philosophy of Science Part C: Studies in
  History and Philosophy of Biological and Biomedical Sciences}, 54:62--71.

\bibitem[{Sun et~al.(2019)Sun, Gaut, Tang, Huang, ElSherief, Zhao, Mirza,
  Belding, Chang, and Wang}]{sun2019mitigating}
Tony Sun, Andrew Gaut, Shirlyn Tang, Yuxin Huang, Mai ElSherief, Jieyu Zhao,
  Diba Mirza, Elizabeth Belding, Kai-Wei Chang, and William~Yang Wang. 2019.
\newblock \href {https://doi.org/10.18653/v1/P19-1159} {{Mitigating Gender Bias
  in Natural Language Processing: Literature Review}}.
\newblock In \emph{Proceedings of the 57th Annual Meeting of the Association
  for Computational Linguistics}, pages 1630--1640, Florence, Italy.
  Association for Computational Linguistics.

\bibitem[{Tan et~al.(2020)Tan, Joty, Kan, and Socher}]{tan-etal-2020-morphin}
Samson Tan, Shafiq Joty, Min-Yen Kan, and Richard Socher. 2020.
\newblock \href {https://doi.org/10.18653/v1/2020.acl-main.263} {{It{'}s
  Morphin{'} Time! {C}ombating Linguistic Discrimination with Inflectional
  Perturbations}}.
\newblock In \emph{Proceedings of the 58th Annual Meeting of the Association
  for Computational Linguistics}, pages 2920--2935, Online. Association for
  Computational Linguistics.

\bibitem[{Tjong Kim~Sang and
  De~Meulder(2003)}]{tjong-kim-sang-de-meulder-2003-introduction}
Erik~F. Tjong Kim~Sang and Fien De~Meulder. 2003.
\newblock \href {https://www.aclweb.org/anthology/W03-0419} {{Introduction to
  the {C}o{NLL}-2003 Shared Task: Language-Independent Named Entity
  Recognition}}.
\newblock In \emph{Proceedings of the Seventh Conference on Natural Language
  Learning at {HLT}-{NAACL} 2003}, pages 142--147.

\bibitem[{Webster et~al.(2018)Webster, Recasens, Axelrod, and
  Baldridge}]{webster-2018}
Kellie Webster, Marta Recasens, Vera Axelrod, and Jason Baldridge. 2018.
\newblock \href {https://doi.org/10.1162/tacl\_a\_00240} {{Mind the GAP: A
  Balanced Corpus of Gendered Ambiguous Pronouns}}.
\newblock \emph{Transactions of the Association for Computational Linguistics},
  6:605--617.

\bibitem[{Zhao et~al.(2018)Zhao, Wang, Yatskar, Ordonez, and
  Chang}]{zhao-etal-2018-gender}
Jieyu Zhao, Tianlu Wang, Mark Yatskar, Vicente Ordonez, and Kai-Wei Chang.
  2018.
\newblock \href {https://doi.org/10.18653/v1/N18-2003} {{Gender Bias in
  Coreference Resolution: Evaluation and Debiasing Methods}}.
\newblock In \emph{Proceedings of the 2018 Conference of the North {A}merican
  Chapter of the Association for Computational Linguistics: Human Language
  Technologies, Volume 2 (Short Papers)}, pages 15--20, New Orleans, Louisiana.
  Association for Computational Linguistics.

\bibitem[{Zhiltsova et~al.(2019)Zhiltsova, Caton, and
  Mulway}]{Zhiltsova2019MitigationOU}
A.~Zhiltsova, S.~Caton, and Catherine Mulway. 2019.
\newblock \href {http://ceur-ws.org/Vol-2563/aics_30.pdf} {{Mitigation of
  Unintended Biases against Non-Native English Texts in Sentiment Analysis}}.
\newblock In \emph{Proceedings for the 27th AIAI Irish Conference on Artificial
  Intelligence and Cognitive Science}.

\end{thebibliography}
\bibliographystyle{acl_natbib}

\end{document}